\documentclass[10pt,journal,compsoc]{IEEEtran}
\usepackage{multirow}
\usepackage{xcolor}
\usepackage{adjustbox}
\definecolor{blue}{rgb}{0,0.0,1.0}
\definecolor{red}{rgb}{1.0,0,0.}
\definecolor{green}{rgb}{0.0,1.0,0.}
\usepackage{graphicx}
\usepackage{hyperref}
\hypersetup{
    colorlinks=true,
    linkcolor=blue,
    filecolor=magenta,      
    urlcolor=cyan,
    pdftitle={Overleaf Example},
    pdfpagemode=FullScreen,
    }
\usepackage{caption}  
\usepackage[ruled]{algorithm2e}

\ifCLASSOPTIONcompsoc
  \usepackage[nocompress]{cite}
\else
  \usepackage{cite}
\fi
\ifCLASSINFOpdf
\else
\fi
\begin{document}
\title{A Semantic, Syntactic, And Context-Aware Natural Language Adversarial Example Generator}


\author{Javad~Rafiei Asl,
        Mohammad H.~Rafiei,
        Manar~Alohaly,
        and~Daniel~Takabi% <-this % stops a space
\IEEEcompsocitemizethanks{\IEEEcompsocthanksitem J. Rafiei Asl is with the Department of Computer Science, Georgia State University, Atlanta,
GA, 30303. E-mail: jasl1@student.gsu.edu
\IEEEcompsocthanksitem M. H. Rafiei is with the Department of Computer Science, Georgia State University, Atlanta,
GA, 30303. E-mail: mrafiei@gsu.edu
\IEEEcompsocthanksitem M. Alohaly is with the Department of Information Systems, College of Computer and Information Sciences, Princess Nourah Bint Abdulrahman University, Riyadh, Saudi Arabia . E-mail: mfalohaly@pnu.edu.sa
\IEEEcompsocthanksitem D. Takabi is with the Department of Computer Science, Georgia State University, Atlanta,
GA, 30303. E-mail: takabi@gsu.edu}% <-this % stops an unwanted space
\thanks{}}


\markboth{IEEE TRANSACTIONS ON DEPENDABLE AND SECURE COMPUTING}%
{Shell \MakeLowercase{\textit{et al.}}: Bare Demo of IEEEtran.cls for Computer Society Journals}



\IEEEtitleabstractindextext{%
\begin{abstract}
Machine learning models are vulnerable to maliciously crafted Adversarial Examples (AEs). Training a machine learning model with AEs improves its robustness and stability against adversarial attacks. It is essential to develop models that produce high-quality AEs. Developing such models has been much slower in natural language processing (NLP) than in areas such as computer vision. This paper introduces a practical and efficient adversarial attack model called SSCAE for \textbf{S}emantic, \textbf{S}yntactic, and \textbf{C}ontext-aware natural language \textbf{AE}s generator. SSCAE identifies important words and uses a masked language model to generate an early set of substitutions. Next, two well-known language models are employed to evaluate the initial set in terms of semantic and syntactic characteristics. We introduce (1) a dynamic threshold to capture more efficient perturbations and (2) a local greedy search to generate high-quality AEs. As a black-box method, SSCAE generates humanly imperceptible and context-aware AEs that preserve semantic consistency and the source language's syntactical and grammatical requirements. The effectiveness and superiority of the proposed SSCAE model are illustrated with fifteen comparative experiments and extensive sensitivity analysis for parameter optimization. SSCAE outperforms the existing models in all experiments while maintaining a higher semantic consistency with a lower query number and a comparable perturbation rate.   
\end{abstract}

}

\maketitle

\IEEEdisplaynontitleabstractindextext

% \fi
\IEEEpeerreviewmaketitle

\IEEEraisesectionheading{\section{Introduction}\label{sec:introduction}}

\IEEEPARstart{D}{espite} the machine learning superiority in various computer science domains, their vulnerability to \textbf{Adversarial Example}s (\textbf{AE}s), i.e., maliciously crafted perturbations, remains an open area of research \cite{goodfellow2014explaining, kurakin2016adversarial, zhang2020adversarial}. The implication of these humanly imperceptible perturbations is to force a fine-tuned machine learning model (dubbed a target model) to produce wrong decisions that align with attackers' intentions. Recent research shows that introducing AEs to a target model during the training, referred to as adversarial training, improves the robustness and stability of that model against adversarial attacks \cite{shafahi2020universal, xu2020adversarial, wang2021convergence}. It would help develop computational models that generate well-design humanly imperceptible adversarial attack examples for adversarial training.

While several adversarial attack and defense studies have contributed to machine vision \cite{goodfellow2014explaining, papernot2017practical, chakraborty2018adversarial}, the progress of this problem in natural language processing (NLP) has been at a much slower pace. Designing practical adversarial attack and defense techniques in NLP is more challenging due to the discrete nature of the text \cite{jin2020bert}. A well-crafted perturbation in NLP fools the target model while establishing three essential principles: (1) compatibility with the human decision \cite{jin2020bert, li2020bert}, (2) preserving the semantic consistency of the original text \cite{jin2020bert, song2020universal}, and (3) following the source language's syntactic and grammatical rules \cite{jin2020bert, song2020universal}.

\begin{table*}[!t]
\caption{Qualitative examples of adversarial samples generated by different adversarial attack methods (TextFooler, BAE, SSCAE). We only attack premises in the MNLI task.}
\label{tbl118}
\centering
\begin{center}
\renewcommand{\arraystretch}{1.6}
\begin{tabular}{ c c p{9cm} c}
\hline
\bfseries Dataset & \bfseries Adversarial Attack  & \bfseries Text Instance & \bfseries Label \\
\hline\hline

\multirow{4}{*}{YELP} & Original & Great location! Close to shops and theatre. Nice staff. & Positive \\
& {TextFooler} & \textbf{Poor} location! \textbf{Distance} from shops and theatre. \textbf{Unpleasant} staff. & Negative   \\
& {BAE} & \textbf{Excellent} location! \textbf{Incredibly} \textbf{near} to shops and theatre. Nice staff. & Negative   \\
& {SSCAE (ours)} & \textbf{Terrible} location! Close to shops and theatre. Nice staff. & Negative  \\

\hline

\multirow{7}{*}{MNLI} & Original & Hypothesis: Poirot was disappointed with me $ \rightarrow$ Premise: Still, it would be interesting to know. 109 Poirot looked at me very earnestly, and again shook his head. &  Neutral   \\
& {TextFooler} & Still, it would be \textbf{enthusiastic} to know. 109 Poirot looked at me very \textbf{cautious}, and again \textbf{vibrate} his head. & Entailment   \\
& {BAE} & Still, it would be \textbf{captivating} to know. 109 Poirot \textbf{gazed} at me \textbf{extremely} earnestly, and again shook his head & Entailment   \\
& {SSCAE (ours)} & Still, it would be interesting to know. 109 Poirot looked at me very \textbf{carefully}, and again shook his head. &  Entailment  \\

\hline

\multirow{5}{*}{IMDB} & Original & You'd better choose Paul Verhoeven's even if you have watched it. & Negative   \\
& {TextFooler} & You'd \textbf{brilliantly } choose Paul Verhoeven's even if you have \textbf{seen} it. & Positive   \\
& {BAE} & You'd \textbf{certainly} choose Paul Verhoeven's even if you \textbf{haven't} watched it. & Positive   \\
& {SSCAE (ours)} & You'd \textbf{rather} choose Paul Verhoeven's even if you have watched it. & Positive  \\

\hline

\end{tabular}
\end{center}
\end{table*}

TextFooler \cite{jin2020bert}, one of the popular adversarial attack models in NLP, first employs a word embeddings technique to explore potential synonyms for each important word. Next, it applies grammatical and semantic similarity checks to narrow down the synonyms and find proper substitutions to serve as perturbations. However, some studies showed that TextFooler produces complicated out-of-context and human-detectable perturbations \cite{garg2020bae, li2020bert}. To address this problem, two recent adversarial attack models, Bidirectional Encoder Representations from Transformers (BERT) Attack (BERT-Attack) \cite{li2020bert} and BERT-based AEs (BAE) \cite{garg2020bae}, were proposed to generate contextual perturbations. The idea is to mask and replace important words with substitutions produced by BERT Masked Language Model (BERT MLM) \cite{devlin2018bert}. Although the generated perturbations seem context-aware substitutions, the original text's semantic and syntactic aspects are sometimes lost \cite{li2020bert}, jeopardizing the robustness and effectiveness of BERT-Attack and BAE. There is a need for a comprehensive model that simultaneously meets all principles mentioned above for well-crafted perturbations.

This paper introduces a practical and efficient adversarial attack model referred to as SSCAE for \textbf{S}emantic (principle 2), \textbf{S}yntactic (principle 3), and \textbf{C}ontext-aware (principle 1) natural language \textbf{AE}s generator. SSCAE is a black-box attack method that generates humanly imperceptible and context-aware AEs preserving semantic consistency and the source language's syntactical and grammatical requirements. It first employs the BERT MLM to generate an initial ``set'' of substitution candidates. Next, it applies two well-known language models, Universal Sentence Encoder (USE) \cite{cer2018universal} and Generative Pre-trained Transformer 2 (GPT-2) \cite{radford2019language} to evaluate the initial set in terms of semantic and syntactic characteristics, respectively. We introduce a new thresholding technique, dubbed dynamic threshold, to meet semantic and syntactic requirements per important word instead of a constant threshold, the focus of recent studies \cite{jin2020bert, li2018textbugger, garg2020bae}. A constant threshold might distinguish substitution's semantic similarity for one word but poorly address the superiority of a different substitution. By independently computing this threshold for every important word, SSCAE, for the first time in the area of adversarial attack, utilizes the non-linearity of the semantic and syntactic correctness distribution to capture more efficient perturbations than static thresholding. Furthemore, the SSCAE model explores the matching of the POS tag (Part-Of-Speech tag) of the substitutions with the original word as a means to uphold grammatical linguistic characteristics. Additionally, we introduce a word permutation-substitution technique, dubbed local greedy search, to substitute several important words simultaneously to generate high-quality AEs. SSCAE uses local greedy search to extract candidate substitutions leading to the most reduction of the confidence score, and produces high-quality AEs using combinations of different substitutions. Table \ref{tbl118} shows qualitative examples of adversarial samples generated by three different adversarial attacks: TextFooler, BAE, and SSCAE methods. In contrast to the other two adversarial attacks, our model generates more natural and high-quality adversarial examples using only a few substitutions, while preserving semantic consistency and the source language's syntactical and grammatical requirements.

We conducted fifteen computational experiments employing frequently used text classification and natural language inference (NLI) datasets to (1) illustrate SSCAE's performance as compared with three state-of-the-art methods: TextFooler, BERT-Attack, and BAE using the BERT target model (seven experiments), and (2) illustrate SSCAE's effectiveness on other target models and different NLP-based cloud applications (eight experiments). SSCAE outperforms the other methods in all experiments while maintaining a higher semantic consistency with a lower query number and a comparable perturbation rate.

\textbf{Contributions:} Our main contributions are summarized as follows:
\begin{itemize}

\item We propose a practical and efficient adversarial attack method, SSCAE, to generate imperceptible context-aware adversarial examples. To do so, our method takes different contextual linguistic requirements (i.e., semantic, syntactic, and grammatical features) into account to refine candidate substitutions suggested by BERT-MLM, resulting in high-quality perturbations. The proposed SSCAE model includes two adroit novelties that are (1) dynamic thresholding and (2) word permutation-substitution to beat state-of-the-art adversarial attack models with a relatively large margin.\footnote{Our code are publicly available at \url{https://github.com/jasl1/SSCAE}} 

\item  We introduce a dynamic threshold technique to meet semantic and syntactic requirements per important word instead of a constant threshold. By independently computing this threshold for every important word, SSCAE, for the first time, considers the non-linearity of the semantic and syntactic correctness distribution. The experimental results confirmed the significance of dynamic thresholding compared with constant thresholding in terms of (1) after-attack accuracy, (2) average perturbation percentage, and (3) average semantic consistency in both text classification and natural language inference tasks.

\item We introduce a word permutation-substitution technique, dubbed local greedy search, to substitute several important words simultaneously to generate high-quality adversarial examples. The experimental results confirmed the significance of local greedy search compared with conventional sequential substitutions in terms of (1) after-attack accuracy, (2) average query number, and (3) average semantic consistency. In terms of average perturbation percentage, SSCAE outperformed four out of seven different NLP tasks.

\item We evaluate SSCAE on different deep learning models over four frequently used text classification and three natural language inference (NLI) tasks. Our method outperforms the state-of-the-art methods by maintaining a higher after-attack accuracy and semantic consistency with a lower query number and a comparable perturbation rate.

\end{itemize}

\section{Related Work}
This section categorizes adversarial attacks on textual data into two groups: white-box and black-box.

In a black-box setting, the proposed approaches range from character-level to sentence-level techniques; the word-level methods demonstrate their superiority compared to other approaches \cite{jia2017adversarial, li2018textbugger,zhang2020adversarial}. Jia and Liang \cite{jia2017adversarial} proposed concatenative adversaries to append distracting sentences at the end of a paragraph to attack the Stanford Question Answering reading comprehension system. Jin et al. \cite{jin2020bert} introduced TextFooler, identifying the important words, gathering a candidate set of possible synonyms, and replacing each important word with the most semantically similar and grammatically correct synonym. Li et al.\cite{li2020bert} proposed BERT-Attack, which consists of two steps: (1) searching for the vulnerable tokens (word/sub-words) (2) employing BERT MLM to generate semantic-preserving substitutes for the vulnerable tokens. Garg and Ramakrishnan \cite{garg2020bae} proposed the BAE model utilizing four adversarial attack strategies based on word replacing or/and word insert operations in the original text. The BAE model masks a portion of the text and then uses the BERT MLM to generate substitutions. Maheshwary et al. \cite{maheshwary2021generating} proposed a decision-based attack strategy to discover word replacements that maximize the semantic similarity between the original and adversarial text. Liu et al. \cite{liu2023aliasing} presented a novel score-based attack model that addresses the issue of selecting important words in textual attack models. The model generates semantically adversarial examples to deceive text classification models. It incorporates the self-attention mechanism and confidence probabilities to determine the significance of words. Lee et al. \cite{lee2022query} presented a black-box attack method that efficiently queries the target model using Bayesian optimization. Their approach utilizes automatic relevance determination (ARD) categorical kernel to dynamically identify important positions in the input. To enhance scalability for longer input sequences, They introduced block decomposition and history subsampling techniques in Bayesian optimization. Wu et al. \cite{wu2023prada} presented the word substitution ranking attack task, which focuses on enhancing the ranking of a specific document by introducing adversarial perturbations into its text. To accomplish this, They proposed a new method called Pseudo Relevance-based ADversarial ranking Attack (PRADA). PRADA involves training a surrogate model using Pseudo Relevance Feedback to generate gradients that aid in discovering the appropriate adversarial perturbations. Fursov et al. \cite{fursov2022differentiable} introduced a novel black-box attack technique at the sentence level. Their approach involves fine-tuning a pre-trained language model to generate adversarial examples. They proposed a differentiable loss function that relies on the score of a substitute classifier and an approximate edit distance, which is calculated using a deep learning model. Liu et al. \cite{liu2022order} introduced an imitation adversarial attack designed for black-box neural passage ranking models. By utilizing a ranking imitation model, They strategically manipulate the ranking results and transfer this manipulation attack to the target ranking model. They proposed a novel gradient-based attack method to generate adversarial triggers with only a few tokens, leading to intentional disorderliness in the rankings. Lu et al. \cite{lu2023black} introduced two efficient black-box attackers to defeat three target systems: MLP, AutoEncoder, and DeepLog. The first attacker is attention-based, utilizing attention weights to identify vulnerable logkeys and generate adversarial examples. The second attacker is gradient-based, calculating gradients based on potential vulnerable logkeys to find the optimal adversarial sample. Both attackers are effective in compromising the target systems and undermining their security.
 
In contrast to the black-box, the white-box setting provides access to the target model architecture, parameters, and the training dataset. Ebrahimi et al. \cite{ebrahimi2017hotflip} developed HotFlip that benefits from an atomic flip operation to select the best character-level change (from the insert, delete, and swap operations) to produce AEs. Li et al. \cite{li2018textbugger} proposed the TEXTBUGGER framework that works in both white-box and black-box settings. Regarding the white-box setting, TEXTBUGGER identifies important words by computing the Jacobian matrix of the target model and selects an optimal perturbation from five types of generated perturbations. Regarding the black-box setting, the framework first identifies important sentences and then specifies the important words using a scoring function, and finally uses the optimal perturbation selection algorithm to change the selected words. Song et al. \cite{song2020universal} proposed Natural Universal Trigger Search (NUTS). It employs a regularized autoencoder \cite{zhao2018adversarially} to generate adversarial attack triggers. Then a gradient-based search is developed to identify triggers with a good attack performance. Chuan et al. \cite{guo2021gradient} proposed Gradient-based Distributional Attack (GBDA), including two key components: first, AEs are instantiated with the Gumbelsoftmax distribution \cite{jang2016categorical}, second, perceptibility and fluency characteristics are enforced using BERTScore \cite{zhang2019bertscore} and a causal language model perplexity, respectively.

SSCAE is a black-box attack model, unaware of the target model's specification. It only has access to the input and output of the target model via an API call. In other words, it feeds the target model with an input sentence and receives a predicted label and its corresponding confidence score. Compared to similar and recent related work \cite{maheshwary2021generating, liu2023aliasing, li2020bert}, our model introduces (1) a new thresholding technique (dynamic threshold) to refine efficient context-aware perturbations and (2) a word substitution technique (local greedy search) leading to high-quality AEs that fool state-of-the-art target models and achieve the best performance in most evaluation metrics compared with state-of-the-art adversarial attacks such as Bert-Attack \cite{li2020bert}, TextFooler \cite{jin2020bert}, and Liu2023 \cite{liu2023aliasing}. Moreover, the proposed model utilizes a comprehensive set of linguistic refinements to improve the imperceptibility of the output AEs from the human perspective. A human evaluation section verifies the quality of the generated AEs by our proposed model.            

\section{Proposed SSCAE Computational Model}
\begin{figure}[h]
    \centering
    \includegraphics[width=0.48\textwidth]{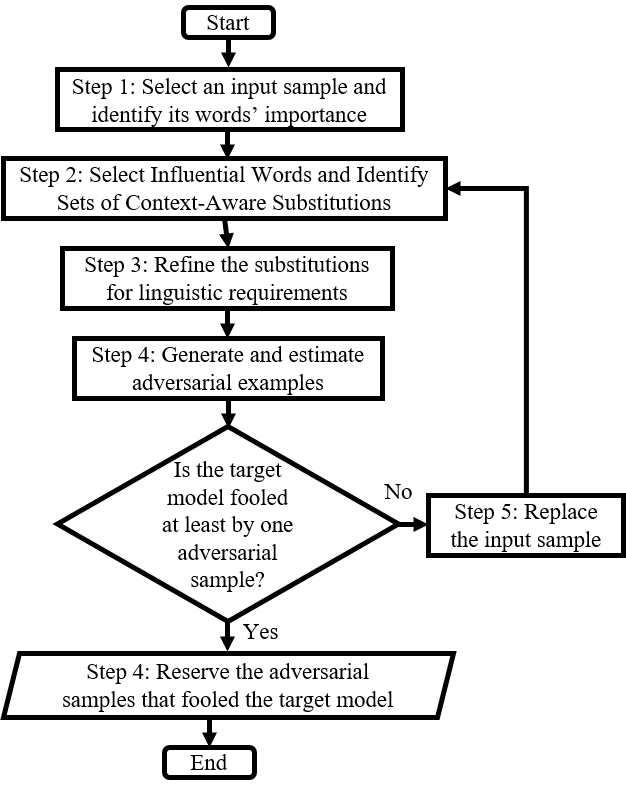}
    \caption{General architecture of the proposed SSCAE model}
    \label{fig1}
\end{figure}

Figure \ref{fig1} presents the flowchart of the proposed SSCAE model. It includes five steps to be explained in this section. A brief background sheds light on two NLP models useful for addressing linguistic constraints before focusing on the proposed model.  

\subsection{Background}
\subsubsection{Universal Sentence Encoder (USE) and Semantic Similarity}
Cer et al. \cite{cer2018universal} developed ``the USE for English'' to encode textual content with arbitrary length into an embedding vector with a predefined size (usually 512) that preserves semantic characteristics of the textual contents statistically. Such embedding vectors are practical for measuring semantic similarities between arbitrary length textual contents \cite{jin2020bert}. The proposed SSCAE model employs USE to generate embedding vectors enriched with semantic characteristics of the input sentence ($V_{input}$) and the AE ($V_{AE}$). Next, the cosine similarities ($S_{similarity}$) between the embedding vectors are computed to verify the semantic similarity of each generated AE to its input sentence:

\begin{equation}
\label{syntactic_correctness}
S_{similarity} = \frac{V_{input}.V_{AE}}{||V_{input}||\:||V_{AE}||}
\end{equation}

\subsubsection{Generative Pre-Trained Transformer 2 (GPT-2)}
Radford et al. \cite{radford2019language} developed GPT-2, a transformer-based language model that computes the probability of a specific word to be the next word in a sentence using previous words. The GPT-2 pre-trained model has been utilized in numerous scientific studies \cite{chen2022unsupervised, pang_etal_2020_towards} to assess the language fluency of the input text based on the syntactic principles of the source language. The proposed SSCAE model employs GPT-2 first to compute the probability of the important word, {$P_{I}$}, and each of its corresponding substitutions, {$P_{S}$}. The syntactic correctness score for a substitution is computed as follows:
\begin{equation}
\label{syntactic_correctness}
S_{syntactic} = P_S - P_I
\end{equation}

\subsection{Step 1: Select an Input Sample and Identify its Words' Importance}
A textual input sample is randomly selected from an available dataset and inputted into the SSCAE model. The input sample can be a sentence, a paragraph, or a document. A greedy search method \cite{gao2018black} is employed to identify the input sample's important words. It masks a word (i.e., replaced with a predefined symbol) in the input sample at a time. Next, the target model evaluates the masked input sample and estimates a confidence score for the truth label. A confidence score is a value between 0 and 1 that represents the likelihood of an input belonging to a label, computed by feeding an input sample (whether masked or not) to a trained model (i.e., target model). It is also referred to as logits in machine learning literature \cite{jordan2015machine, lecun2015deep, mahesh2020machine}. The difference between the confidence score before and after masking, denoted as {$\delta$}, is recorded for that particular word. The value of {$\delta$} is computed for all of the words in the input sample. Ultimately, the words are sorted based on the value of {$\delta$}; the larger the {$\delta$} value, the more influential (important) is the corresponding word in the input sample for the target model. 

%algorithm example url
%https://tex.stackexchange.com/questions/403823/how-to-use-function-in-latex-algorithm
\begin{algorithm*}
    \SetKwInOut{Input}{Input}
    \SetKwInOut{Output}{Output}

    \Input{input sentence {$S=[w_1, ..., w_n]$}, truth label {$Y$}, target model {$T$}, Fixed-size window {$W$}, thresholds' lower bounds ({$ST_{semantic}$}, {$ST_{syntactic}$})}
    \Output{Adversarial Sentence {$S^{adv}$}} 

     \SetKwProg{myproc}{\\Function}{}{}
      \myproc{Compute\_Importance ({$S^{adv}$}, {$T$}, {$Y$}):}{
      \ForEach{word $w_i \in$  $S^{adv}$}{
    Calculate {$p_i(Y)$} of {$ S_{i-masked}^{ adv}$} through {$T$}\\
    }
    \KwRet Sorted words {$WL = [w_{min_1 (P(Y))},w_{min_2 (P(Y))},...]$}
      }
    \textbf{End Function}\\\
    
      \SetKwProg{myproc}{Procedure}{}{}
      \myproc{SSCAE\_Attack ({$S$}, {$M$}, {$N$}, {$T$}, {$Y$}, {$W$}, {$ST_{semantic}$}, {$ST_{syntactic}$}):}{
      Initialization {$S^{adv}$} {$\leftarrow$}  {$S$} \\
      WL = Compute\_Importance({$S^{adv}$}, {$T$}, {$Y$})\\
      {$TEMP\_M = 0$}\\
      {$SELECTED\_WL = []$}\\
      \ForEach{word  $w_i \in$ WL}{
        Apply MLM\_Bert on {$w_{(i \pm j)}$} using {$W$}, gather semantic candidates ({$SC$})\\
        {$S_{semantic} = USE(S^{adv}, SC, w_i)$}\\
        {$S_{syntactic} = GPT\_2(S^{adv}, SC, w_i)$}\\
        Calculate dynamic thresholds ({$DT_{semantic},DT_{syntactic}$}) based on ({$S_{semantic},S_{syntactic}$})\\
        {$PC = []$}  //purified\_candidates \\
        
        \ForEach{semantic candidate  $sc_k \in$ SC}{
            \If{{$ S_{semantic}[sc_k]> (DT_{semantic},ST_{semantic}) $}  AND   {$S_{syntactic}[sc_k] > (DT_{syntactic},ST_{syntactic}) $}}
            {
                \If{$ POS\_checker(w_i, sc_k) == True $}
                {
                    {$PC.add(sc_k)$}
                }
            }
        }
        {$GAP = []$}\\
        \ForEach{purified candidate  $pc_k \in$ PC}{
        {$S_t^{adv}= [...,w_{i-1}, pc_k  ,w_{i+1}  ,...] $}\\
            \If{$argmax(T(S_t^{adv})) \ne Y $}
            {
                \KwRet {$S_t^{adv}$}
            }
            \ElseIf{$T_y (S_t^{adv}) < T_y (S^{adv})$}
            {
                {$GAP.add(T_y (S^{adv})-T_y (S_t^{adv}))$}\\
                
            }
        }

       }
     $TEMP\_M = TEMP\_M + 1$  \\  
     $SELECTED\_WL.add(w_i)$ \\
      Sort $GAP$ descendingly \\
     \If{$TEMP\_M >= M$}
        {
            $TEMP\_M = 0$ \\
            $local\_substitutions = dict()$ \\
            \ForEach{$w_j$  $\in$ SELECTED\_WL}{
                $local\_substitutions[w_j] = PC[GAP[N]]$ \\
            }
            products = [PRODUCT(*local\_substitutions)] \\
            \ForEach{$S_t^{adv}$  $\in$ products}{
            \If{$argmax(T(S_t^{adv})) \ne Y $}
            {
                \KwRet {$S_t^{adv}$}
            }
            }
        }
        
        \KwRet $S^{adv}$ \\\
      }
     \textbf{End Procedure} 
    \caption{SSCAE Pseudocode}
    \label{algorithm1}
\end{algorithm*}

\subsection{Step 2: Select Influential Words and Identify Sets of Context-Aware Substitutions}
{$\mathrm{M}$} words with the largest {$\delta$}s (i.e., most influential) are selected to be substituted. For each important word, a set of neighboring words is selected by utilizing a window of size 2*{$\mathrm{N_W}$}, which is centered on the important word. This window includes {$\mathrm{N_W}$} neighbors before and {$\mathrm{N_W}$} neighbors after the important word. Through trial-and-error, it was determined that a window size of 4 (i.e., {$\mathrm{N_W}$} = 2) is sufficiently large to capture a comprehensive set of context-aware substitutions. Instead of directly applying the BERT MLM on the important word to retrieve candidate substitutions, an alternative approach is to apply the BERT MLM on a neighboring word. This allows for the retrieval of a set of meaningful substitutions specifically for the important word. Considering a fixed-size word window of four words, the SSCAE model applies the BERT MLM on each neighboring word of the important word. This process results in the retrieval of distinct sets (i.e., four sets) of the top {$K$} context-aware substitutions from different aspects of the view. These sets are then combined using a union operation, yielding a combined set that serves as the initial set of context-aware substitutions for the selected influential word. It should be noted that the initial set is larger than {$K$} but close to {$K$}; many substitutions are identical across top {$K$} context-aware sets. This helps retrieve identical and non-identical substitutions to improve the target model's chance of fooling.

\subsection{Step 3: Refine the Substitutions for Linguistic Requirements with Dynamic Threshold}
In order to account for the non-linear distribution of semantic similarity and syntactic correctness in the context-aware substitutions, we propose the implementation of an adaptive and individualized threshold for each significant word. Termed the ``dynamic threshold,'' this threshold is specifically designed to alleviate the constraints imposed by a fixed threshold. It accomplishes this by dynamically computing an appropriate threshold based on linguistic features, namely semantic similarity and syntactic correctness, for the candidate substitutions of each important word. The primary objective of this approach is to generate perturbations that exhibit enhanced quality and fluency. 

Although BERT-based models demonstrate their superiority in various NLP tasks \cite{devlin2018bert, liu2019roberta, lan2019albert}, unfortunately, most top {$K$}-generated substitutions do not eventually lead to valid AEs in terms of language fluency, semantic consistency, and imperceptibility. It is necessary to scrutinize the substitutions to ensure the validity of produced perturbations and comply with linguistic aspects. In other words, each substitution has to preserve the input sample's meaning and grammatical requirements while following the source language's syntactic structures. The SSCAE model employs refinement strategies to identify substitutions with a higher chance of preserving semantic, syntactic, and grammatical linguistic characteristics. 

USE, as a semantic embedding model, and GPT-2, as a transformer-based language model, assign a semantic similarity and a syntactic correctness score for each substitution, respectively. The semantic similarity score represents to what extent a substitution preserves the meaning of the input sample. The syntactic score represents to what extent a substitution preserves the source language syntax principles. Recent studies utilize a constant threshold in two ways: refining candidate substitutions through iterative comparison with a previous adversarial example \cite{jin2020bert}, and filtering substitutions based on semantic consistency with the original text \cite{li2018textbugger, li2020bert}. However, relying solely on a constant threshold is insufficient for effective refinement, especially when comparing the generated adversarial example with the previous one. While a constant threshold may successfully refine a set of high-quality perturbations for an important word, it significantly reduces semantic similarity with the original text for other important words. Consequently, although such an adversarial attack method can generate adversarial examples that fool the target model, the semantic consistency of the final adversarial examples is greatly compromised, making them easily detectable from a human perspective.

This paper investigates four different potential heuristics to compute dynamic thresholds to refine the substitutions: (1) Average\_threshold, (2) Median\_threshold, (3) TopN\_threshold, and (4) Top\_maxes\_distance. Average\_threshold (Median\_threshold) computes the average (median) of substitutions' scores. TopN\_threshold picks the score of the {$N^{th}$} substitution after being descending-sorted where {$N$} is a minor hyper-parameter. Top\_maxes\_distance computes the specific threshold as follows:
\begin{equation}
\label{Top_maxes_distance}
DT = S_N - \lambda.\Delta S_{N},
\end{equation}
where {$S_N$} is the score corresponding to the {$N^{th}$} substitution after being descending-sorted, {$\lambda$} is a minor hyperparameter, and {$\Delta S_{N}$} is the difference between the highest score amongst substitutions and {$S_{N}$}. Overall, such dynamic thresholds lead to substitutions with higher semantic consistency, significantly improving generated AEs' quality. By extensive trial and error experiments on various datasets, a lower bound of 0.7 was optimal for the dynamic threshold to assure the validity of the substitutions.  

The POS tag (Part-Of-Speech tag) of the substitutions must match that of the selected word to preserve the grammatical linguistic characteristics. Otherwise, the substitution is filtered out. One exception is when the substitution is singular (plural), while the selected word's POS tag indicates a plural (singular) substitution. In this case, the substitution is modified accordingly (not filtered). Another exception is verb substitutions: if a verb substitution has the same root as the selected word, it is filtered. It should be noted that, in this method, a correct substitution's POS tag is identified when the substitution is implemented in the input sample. 

\subsection{Step 4: Generate and Estimate AEs with Local Greedy Search}

As an advanced and enhanced approach, the simultaneous substitution of multiple important words is proposed as a more efficient alternative to sequential substitution. This straightforward and pragmatic concept, referred to as ``local greedy search,'' results in worst-case perturbations and effectively expedites the fooling process of the target model, while maintaining the generation of high-quality and imperceptible adversarial examples.   

The local greedy search method is employed by the proposed SSCAE model to generate adversarial samples: For each of {$\mathrm{M}$} important words (i.e., {$\mathrm{M}$} top words with largest {$\delta$}s) (specified in step 2), SSCAE selects top {$\mathrm{N}$} refined substitutions that lead to the most reduction of the confidence score. To obtain potential adversarial samples, the local greedy search is to substitute all {$\mathrm{M}$} important words simultaneously. Both {$\mathrm{M}$} and {$\mathrm{N}$} are selected such that it keeps the number of combinations of different substitutions, i.e., {$\mathrm{N}^\mathrm{M}$}, within the available computation resources. For example, in this paper, magnitudes of 1 to 4 were investigated for both {$\mathrm{M}$} and {$\mathrm{N}$} such that the total number of combinations is a maximum of 256 (when {$\mathrm{N} \: {=} \:4$} and {$\mathrm{M} \: {=} \: 4$}). Due to adaptive refinements of the previous step, the generated adversarial samples have a higher chance of preserving the input sample's semantic, syntactic, and grammatical consistency. The adversarial samples are then estimated by the target model and descendingly sorted based on the following gap magnitude:

\begin{equation}
\label{confidence_gap}
\mathrm{{G_y}^{adv}} = \mathrm{T_y} (S^{orig}) - \mathrm{T_y} (S^{adv}),
\end{equation}

where $\mathrm{T(.)}$ is the target model such $\mathrm{T_y(.)}$ assigns a confidence score corresponding to truth label ${y}$ for a given input sentence; $S^{orig}$ and $S^{adv}$ are input sample and adversarial sample, respectively. ${{G_y}^{adv}}$ represents the gap between the confidence score of the input sample and the adversarial sample for the truth label. The proposed SSCAE generates adversarial samples by considering all {$\mathrm{N}^\mathrm{M}$} possible combinations of substitutions, where $\mathrm{M}$ important words are replaced with one of the combinations each time. Consequently, the SSCAE model evaluates all {$\mathrm{N}^\mathrm{M}$} adversarial samples to identify one that successfully deceives the target model. The local greedy search idea enables our model to identify a successful adversarial example at early stages of permutation combinations, thus reducing the required number of queries for the adversarial attack objective.

\subsection{Step 5: Replace the Input Sample}
Suppose none of the adversarial samples fool the target model. In that case, the adversarial sample with the lowest probability of truth label (highest gap in equation \ref{confidence_gap}) is selected as the new input text sample. Next, steps 2 to 5 are repeated. It should be noted that, in this case, the new input sample generates adversarial samples with a higher probability of fooling the target model than the current input sample. 

The pseudocode of the proposed SSCAE model is summarized in Algorithm~\ref{algorithm1}. In this algorithm, the main function, {\fontfamily{qcr}\selectfont SSCAE\_Attack()}, receives input text, truth label, target model, a fixed-size window, hyper-parameters for local greedy search, and lower bounds for semantic and syntactic requirements. It returns the ultimate adversarial example. The {\fontfamily{qcr}\selectfont Compute\_Importance()} function calculates the importance score of the words in the input text and sorts them based on their importance. The {\fontfamily{qcr}\selectfont USE()/GPT\_2()} function assigns a semantic/syntactic score to each candidate substitution by replacing the relevant important word with the substitution. The {\fontfamily{qcr}\selectfont POS\_checker()} function checks the POS tag of a substitution and matches it with the important word when the substitution is applied to the input sample. The {\fontfamily{qcr}\selectfont PRODUCT()} function generates all combinations of different substitutions for {$\mathrm{M}$} important words by implementing them within the input sample.

%\IEEEpubidadjcol

%\textcolor{blue}{\ul{We}}
% \textcolor{red}{\st{Then}}

\begin{table*}[!t]
\renewcommand{\arraystretch}{1.2}
\caption{Average results of the proposed SSCAE, TextFooler, BERT-Attack, and BAE adversarial attack models on 1000 randomly selected testing instances from each of seven datasets using the BERT target model. The standard deviation is included within the brackets. (E\#: Experiment Number; Perturb \%: Perturbation Percentage; Query \#: Query Number; H: Hypothesis; P: Premise.)}

\label{tbl1}
\centering
\begin{center}
\begin{tabular}{c c c c c c c c}
\hline
\bfseries Dataset & \bfseries E\# & \bfseries Original Acc & \bfseries Adversarial Attack & \bfseries Attacked Acc & \bfseries Perturb \% & \bfseries Query \# & \bfseries Semantic Sim\\
\hline\hline

\multirow{5}{*}{YELP} & \multirow{5}{*}{1} & \multirow{5}{*}{95.6} & TextFooler & 6.3 (0.5) & 12.1 (0.8) & 758 (32) & 0.75 (0.03) \\
& & & BERT-Attack/BAE & 5.1 (0.4) & \textbf{4.1 (0.3)} & 273 (19) & 0.77 (0.03)\\
& & & HardLabel & 5.6 (0.5) & 5.4 (0.5) & 247 (18) & 0.79 (0.04)\\
& & & Liu2023 & 5.2 (0.3) & 11.5 (0.9) & 294 (21) & 0.80 (0.03)\\
& & & SSCAE (ours) & \textbf{1.6 (0.2)} & 4.8 (0.3) & \textbf{112 (14)} & \textbf{0.92 (0.03)} \\

\hline

\multirow{5}{*}{IMDB} & \multirow{5}{*}{2} & \multirow{5}{*}{90.9} & TextFooler & 13.4 (0.7) & 6.5 (0.6) & 1145 (52) & 0.85 (0.04) \\
& & & BERT-Attack/BAE & 11.4 (0.5) & \textbf{4.4 (0.4)} & 454 (31) & 0.86 (0.03) \\
& & & HardLabel & 10.8 (0.6) & 4.6 (0.5) & 521 (38) & 0.85 (0.03)\\
& & & Liu2023 & 9.5 (0.4) & 6.9 (0.6) & 562 (36) & 0.86 (0.04)\\
& & & SSCAE (ours) & \textbf{4.2 (0.3)} & 10.8 (0.5) & \textbf{418 (26)} & \textbf{0.86 (0.03)} \\

\hline

\multirow{5}{*}{SST2} & \multirow{5}{*}{3} & \multirow{5}{*}{93.0} & TextFooler & 14.1 (0.3) & 16.7 (1.0) & 103 (9) & 0.85 (0.05) \\
& & & BERT-Attack/BAE & 18.2 (0.4) & 14.5 (0.3) & 92 (8) & 0.86 (0.04) \\
& & & HardLabel & 15.4 (0.3) & 14.9 (0.4) & 86 (7) & 0.85 (0.03)\\
& & & Liu2023 & 14.6 (0.3) & 17.1 (0.5) & 90 (6) & 0.85 (0.04)\\
& & & SSCAE (ours) & \textbf{12.5 (0.2)} & \textbf{14.3 (0.3)} & \textbf{64 (5)} & \textbf{0.87 (0.03)} \\
\hline

\multirow{5}{*}{MR} & \multirow{5}{*}{4} & \multirow{5}{*}{85.3} & TextFooler & 24.8 (0.4) & 15.7 (0.3) & 173 (9) & 0.90 (0.03)\\
& & & BERT-Attack/BAE & 19.2 (0.3) & 15.2 (0.2) & 126 (8) & 0.91 (0.02)\\
& & & HardLabel & 18.4 (0.3) & \textbf{15.0 (0.3)} & 119 (7) & 0.88 (0.03)\\
& & & Liu2023 & 18.1 (0.2) & 17.3 (0.2) & 135 (6) & 0.89 (0.02)\\
& & & SSCAE (ours) & \textbf{15.6 (0.2)} & 17.4 (0.2) & \textbf{54 (4)} & \textbf{0.91 (0.02)}\\

\hline

\multirow{5}{*}{SNLI} & \multirow{5}{*}{5} & \multirow{5}{*}{89.4 (H)} & TextFooler & 16.9 (0.3) & 18.0 (0.5) & 74 (6) & 0.73 (0.02) \\
& & & BERT-Attack/BAE & 21.4 (0.4) & 18.8 (0.5) & 26 (3) & 0.71 (0.02) \\
& & & HardLabel & 16.3 (0.4) & 16.9 (0.4) & 37 (4) & 0.71 (0.03)\\
& & & Liu2023 & 15.7 (0.4) & 18.5 (0.6) & 41 (5) & 0.72 (0.03)\\
& & & SSCAE (ours) & \textbf{11.5 (0.3)} & \textbf{16.6 (0.4)} & \textbf{26 (3)} & \textbf{0.76 (0.02)} \\
\hline

\multirow{5}{*}{MNLI-Matched} & \multirow{5}{*}{6} & \multirow{5}{*}{85.1 (P)} & TextFooler & 31.4 (0.6) & 26.5 (0.5) & 232 (10) & 0.76 (0.03) \\
& & & BERT-Attack/BAE & 18.9 (0.4) & 14.5 (0.3) & 64 (4) & 0.78 (0.02) \\
& & & HardLabel & 20.9 (0.5) & 13.6 (0.4) & 114 (6) & 0.75 (0.02)\\
& & & Liu2023 & 18.2 (0.3) & 24.4 (0.5) & 127 (7) & 0.77 (0.02)\\
& & & SSCAE (ours) & \textbf{9.2 (0.3)} & \textbf{12.3 (0.3)} & \textbf{42 (3)} & \textbf{0.82 (0.02)} \\
\hline

\multirow{5}{*}{MNLI-Mismatched} & \multirow{5}{*}{7} & \multirow{5}{*}{82.1 (P)} & TextFooler & 26.1 (0.5) & 27.4 (0.6) & 186 (8) & 0.75 (0.03) \\
& & & BERT-Attack/BAE & 20.7 (0.4) & 15.1 (0.3) & 61 (4) & 0.77 (0.02) \\
& & & HardLabel & 19.7 (0.4) & 14.6 (0.4) & 77 (5) & 0.76 (0.03)\\
& & & Liu2023 & 18.5 (0.5) & 25.7 (0.6) & 81 (6) & 0.78 (0.03)\\
& & & SSCAE (ours) & \textbf{9.6 (0.3)} & \textbf{13.9 (0.3)} & \textbf{35 (3)} & \textbf{0.82 (0.02)} \\
\hline

\end{tabular}
\end{center}
\end{table*}

\section{Computational Experiments}
This section introduces different text classification and NLI datasets and the target models fine-tuned on some (or all) of the datasets. To reduce the statistical bias, each experiment was conducted five times, each time using a different randomly generated set of test samples; the reported results shown in this section are the average results of five repetitions. We employ a set of metrics to evaluate the adversarial attack results of the proposed SSCAE on the different target models compared to recent adversarial attack models. Furthermore, the efficacy of the SSCAE model is examined on various Machine-Learning-as-a-Service (MLaaS) platforms, which includes diverse natural language understanding tasks such as text classification. 

We utilized various libraries for implementing our proposed model, including PyTorch \cite{NEURIPS2019_9015}, spaCy \cite{spacy2}, and Gensim \cite{rehurek2011gensim}. To load different target models and datasets for adversarial purposes, we employed the transformers library \cite{wolf_etal_2020_transformers}. Additionally, we employed the TextAttack library \cite{morris2020textattack} to implement other adversarial attack methods for comparison. Our computational experiments were conducted on four Tesla V100, 32GB GPUs. On average, the SSCAE model took 2-4 hours to run on 1000 randomly selected testing instances.

\subsection{Datasets and Target Models}
To illustrate the efficiency of the SSCAE model, four binary text classification (for text classification task) and three NLI datasets (for text entailment task) were employed to develop eleven NLP task experiments. The text classification datasets are YELP Polarity Review (YELP) \cite{zhang2015character}, Internet Movie Database (IMDb) Review \cite{IMDb}, Rotten Tomatoes Movie Reviews (MR) \cite{pang2005seeing}, and Stanford Sentiment Treebank Version 2 (SST2) \cite{socher2013recursive}. The NLI datasets are Standford NLI (SNLI) \cite{bowman2015large} and two Multi-NLI (MNLI) datasets \cite{williams2017broad}, referred to as MNLI-Matched and MNLI-Mismatched. Details of the datasets are available in Appendix A. The first seven experiments compare the SSCAE model with three state-of-the-art adversarial attack models where the BERT model \cite{devlin2018bert} is employed as the target model and a different dataset is implemented each time for fine-tuning. BERT is a popular transformer-based language model pre-trained on extremely large-scale unlabeled textual data. Recent studies have demonstrated the superiority of the BERT-based models on most NLP tasks; in certain NLP tasks, it has achieved human-level accuracies \cite{Google_BERT, SQuAD}.

The next four experiments aim to demonstrate the effectiveness of the SSCAE model by comparing it with two state-of-the-art adversarial attack models across four neural network target models other than BERT that are WordLSTM \cite{hochreiter1997long}, ALBERT-Base \cite{lan2019albert}, ESIM \cite{chen2016enhanced}, and BERT-Large \cite{devlin2018bert}. WordLSTM addresses the problem of short-term memory in recurrent neural networks by using specific gates to regulate the flow of word-based sequential information. ALBERT utilizes two factorized embedding parameterization and cross-layer parameter sharing to lower the BERT’s memory consumption and increase its training speed. ESIM is a sequential model that enhances the local inference information (words and their context) by calculating the sentence pair’s difference and element-wise product. BERT-Large is a transformer-based model pretrained on a large corpus of English data with 24 layers of encoders stacked on top of each other with 16 bidirectional self-attention heads.   

The last four experiments assess the effectiveness of the SSCAE model on three commercial platforms, i.e., Google Cloud NLP, IBM Watson Natural Language Understanding (IBM
Watson), and Microsoft Azure Text Analytics (Microsoft Azure), and compare its performance with the TEXTBUGGER model \cite{li2018textbugger} to verify its efficiency. These MLaaS platforms deploy machine learning models on cloud servers, allowing users to access the models through an API. The attacker, in such scenarios, lacks knowledge of the model's architecture, parameters, and training data, and can solely query the target model to obtain prediction or confidence score outputs.

\subsection{Comparison of Adversarial Attack Models Against BERT}
Table~\ref{tbl1} presents the results of the first seven experiments. The results compare the proposed SSCAE model with TextFooler, BERT-Attack, BAE, HardLabel \cite{maheshwary2021generating}, and Liu2023 \cite{liu2023aliasing} adversarial attack models using 1000 randomly selected testing instances and the BERT target model. Four standard metrics \cite{jin2020bert, li2020bert} were used to verify the quality of the generated AEs in Table~\ref{tbl1}: (1) after-attack accuracy, (2) average perturbation percentage, (3) average query number, and (4) average semantic consistency. An ideal adversarial attack model would obtain a lower after-attack accuracy, perturbation percentage, and query number and a higher semantic consistency. Due to the similarity and dataset sensitivity of BERT-Attack and BAE models \cite{li2020bert, garg2020bae}, the best literature-available results across these two models are reported in Table~\ref{tbl1} (denoted as BERT-Attack/BAE). Overall, in most datasets, SSCAE outperforms TextFloor, BERT-Attack/BAE, HardLabel, and Liu2023 models, as shown in Table~\ref{tbl1}. 

In the case of after-attack accuracy, SSCAE results are lower (i.e., better), mostly with a large margin, than all other adversarial attack models, particularly in experiments corresponding to YELP, IMDB, SNLI, and MNLIs. In the case of perturbation percentage, SSCAE achieved lower (i.e., better) results than TextFooler and Liu2023 except for IMDB and MR . In the case of MNLI, the perturbation percentage is significantly lower than the TextFooler and Liu2023. Moreover, compared with the Bert-Attack/BAE and HardLabel, SSCAE had lower and comparable perturbation percentages in NLI and text classification (except IMDB) tasks, respectively. In the case of query number, the SSCAE results are significantly lower (i.e., better) than other adversarial attack models in all experiments. In the case of semantic consistency, except for the experiment corresponding to the IMDB and MR datasets, where the results are near-equal, the SSCAE results are always better than other adversarial attack models. One of the promising outcomes of the SSCAE model is that it resulted in the best (i.e., lowest) after-attack accuracy across all experiments while, except for the experiment corresponding to the SNLI dataset, keeping semantic consistency over 0.8, which is exceptional compared with similar studies \cite{jin2020bert, li2020bert}. The proposed SSCAE model demonstrates superior performance compared to other state-of-the-art methods in text classification datasets, exhibiting higher semantic similarity, lower after-attack accuracy and query numbers, and a comparable perturbation rate. This superiority is particularly pronounced in semantic similarity and after-attack accuracy metrics for the YELP dataset. It can be attributed to the successful implementation of key innovations, including the dynamic threshold concept for semantic consistency and syntactic correctness. This idea results in high-quality and fluent adversarial examples, leading to higher (i.e., better) semantic similarity. Additionally, the SSCAE model utilizes local greedy search to simultaneously replace multiple important words, resulting in worst-case perturbations and achieving lower after-attack accuracy. It should be noted that although the SSCAE model's perturbation percentage, as compared with that of BERT-Attack/BAE and HardLabel, is higher (i.e., worse) in three (IMDB and MR) out of seven experiments, its semantic consistency (i.e., imperceptibility) is still preserved over 0.8 across most experiments. This is due to the adroit implementation of semantic, syntactic, and grammatical refinements in the proposed SSCAE model. Although the SSCAE model applies slightly more perturbations in some experiments, it preserves a higher semantic consistency using the dynamic threshold technique. Hence, the SSCAE model generates more imperceptible and efficient AEs than previous state-of-the-art models.

Text entailment tasks (i.e., experiments 5 to 7) are more challenging than the text classification tasks (i.e., experiments 1 to 4) due to their datasets being made of shorter sentences (a few words). Hence, replacing important words would significantly increase the perturbation percentage and reduce the semantic consistency of the generated examples; modifying even one or two important words increases the perturbation percentage, and there are not enough potential words to be perturbed. Nevertheless, as another promising outcome, the SSCAE model remarkably outperforms other state-of-the-art models with a large after-attack accuracy margin on text alignment experiments while achieving a higher semantic consistency. It should be noted that in document-level classification experiments, i.e., YELP and IMDB experiments here, where the length of input samples is relatively larger than that of other experiments, the lower perturbation percentage indicates that the target model, i.e., BERT, relies on only a few important words to make predictions. Therefore, identifying and replacing the important words could reveal the vulnerability of the Bert-base target models \cite{li2020bert}.

\begin{table*}[!t]
\renewcommand{\arraystretch}{1.2}
\caption{Average results of the proposed SSCAE, TextFooler, and BERT-Attack adversarial attack models on 1000 randomly selected testing instances using WordLSTM (with YELP dataset), ALBERT-Base (with YELP dataset), ESIM (with MNLI mismatched), and BERT-Large (with MNLI mismatched) as the target models. The standard deviation is included within the brackets. (E\#: Experiment Number;  Perturb \%: Perturbation Percentage) }

\label{tbl2}
\centering
\begin{center}
\begin{tabular}{c c c c c c c c}
\hline
\bfseries Dataset & \bfseries E\# & \bfseries Target Model & \bfseries Original Acc & \bfseries Attacked Acc & \bfseries Adversarial Attack & \bfseries Perturb \%  & \bfseries Semantic Sim\\
\hline\hline

\multirow{6}{*}{YELP} & \multirow{3}{*}{8} & \multirow{3}{*}{WordLSTM} & \multirow{3}{*}{96.0} & TextFooler & 2.3 (0.2) & 10.8 (0.6) & 0.77 (0.03)\\
 & & & & BERT-Attack & 1.7 (0.1) & 5.5 (0.4) & 0.80 (0.04)\\
 & & & & SSCAE (ours) & \textbf{0.56 (0.05)} & \textbf{5.3 (0.3)} & \textbf{0.93 (0.02)}\\
 \cline{2-8}
 & \multirow{3}{*}{9} & \multirow{3}{*}{ALBERT-Base} & \multirow{3}{*}{97.0} & TextFooler & 6.9 (0.3) & 11.7 (0.4) & 0.74 (0.02)\\
 & & & & BERT-Attack & 5.4 (0.2) & \textbf{4.3 (0.3)} & 0.78 (0.03) \\
 & & & & SSCAE (ours) & \textbf{1.9 (0.1)} & 4.5 (0.2) & \textbf{0.91 (0.02)} \\
\hline

\multirow{6}{*}{MNLI-Mismatched} & \multirow{3}{*}{10} & \multirow{3}{*}{ESIM} & \multirow{3}{*}{76.2} & TextFooler & 16.6 (0.6) & 25.9 (0.7) & 0.74 (0.03)\\
 & & & & BERT-Attack & 10.5 (0.3) & 20.2 (0.5) & 0.77 (0.02)\\
 & & & & SSCAE (ours) & \textbf{6.1 (0.3)} & \textbf{17.5 (0.4)} & \textbf{0.83 (0.03)}\\
 \cline{2-8}
 & \multirow{3}{*}{11} & \multirow{3}{*}{BERT-Large} & \multirow{3}{*}{86.4} & TextFooler & 26.4 (0.5) & 26.8 (0.6) & 0.73 (0.03)\\
 & & & & BERT-Attack & 20.3 (0.4) & 14.9 (0.3) & 0.76 (0.03) \\
 & & & & SSCAE (ours) & \textbf{9.9 (0.3)} & \textbf{13.6 (0.3)} & \textbf{0.81 (0.02)} \\
\hline

\end{tabular}
\end{center}
\end{table*}

\begin{table*}[!t]
\renewcommand{\arraystretch}{1.2}
\caption{Average results of the proposed SSCAE and TEXTBUGGER adversarial attack models on 500
randomly selected testing instances from the IMDB and YELP datasets using different NLP-based cloud APIs. The standard deviation is included within the brackets. (E\#: Experiment Number;  Perturb \%: Perturbation Percentage)}

\label{tbl9}
\centering
\begin{center}
\begin{tabular}{c c c c c c c c}
\hline
\bfseries Dataset & \bfseries E\# & \bfseries Adversarial Attack & \bfseries Original Acc & \bfseries Targeted Model & \bfseries Attacked Acc & \bfseries Perturb \% & \bfseries Semantic Sim\\
\hline\hline

\multirow{6}{*}{IMDB} &
\multirow{3}{*}{12} &
\multirow{3}{*}{TEXTBUGGER} &
86.2 & Google Cloud NLP & 23.3 (1.4) & \textbf{2.6 (0.2)} & 0.72 (0.04) \\
& & & 91.8 & Microsoft Azure & 2.1 (0.3) & {6.4 (0.4)} & 0.70 (0.03)\\
& & & 91.4 & IBM Watson & {4.7 (0.4)} & 8.9 (0.5) & {0.69 (0.04)} \\

\cline{2-8}

& \multirow{3}{*}{13} & 
\multirow{3}{*}{SSCAE (ours)} &
86.2 & Google Cloud NLP & \textbf{17.6 (0.9)} & \textbf{2.6 (0.2)} & \textbf{0.87 (0.03)} \\
& & & 91.8 & Microsoft Azure & \textbf{1.6 (0.2)} & \textbf{6.0 (0.4)} & \textbf{0.85 (0.02)}\\
& & & 91.4 & IBM Watson & \textbf{2.8 (0.3)} & \textbf{7.5 (0.4)} & \textbf{0.85 (0.03)} \\

\hline

\multirow{6}{*}{YELP} &
\multirow{3}{*}{14} &
\multirow{3}{*}{TEXTBUGGER} &
90.6 & Google Cloud NLP & 8.3 (0.7) & \textbf{4.5 (0.3)} & 0.75 (0.03) \\
& & & 85.2 & Microsoft Azure & 3.9 (0.3) & {5.1 (0.3)} & 0.73 (0.04)\\
& & & 86.5 & IBM Watson & {4.6 (0.3)} & 5.9 (0.4) & {0.72 (0.03)} \\

\cline{2-8}

& \multirow{3}{*}{15} & 
\multirow{3}{*}{SSCAE (ours)} &
90.6 & Google Cloud NLP & \textbf{5.6 (0.5)} & 4.7 (0.3) & \textbf{0.91 (0.03)} \\
& & & 85.2 & Microsoft Azure & \textbf{2.1 (0.2)} & \textbf{5.0 (0.3)} & \textbf{0.89 (0.03)}\\
& & & 86.5 & IBM Watson & \textbf{2.8 (0.2)} & \textbf{5.6 (0.4)} & \textbf{0.87 (0.02)} \\
\hline

\end{tabular}
\end{center}
\end{table*}

\subsection{Adversarial Attack Against Target Models Other Than BERT}
Table~\ref{tbl2} presents the experimental results of the proposed SSCAE, TextFooler, and BERT-Attack adversarial attack models on 1000 randomly estimated testing instances using WordLSTM (with YELP dataset), ALBERT-Base (with YELP dataset), ESIM (with MNLI-Mismatched), and BERT-Large (with MNLI-Mismatched) as the target models. In all the conducted experiments, the SSCAE model demonstrates superior performance compared to the TextFloor and BERT-Attack models, exhibiting lower after-attack accuracy, higher semantic consistency, and a comparable perturbation percentage. In the case of WordLSTM-YELP and ALBERT-Base-YELP (i.e., text classification tasks) experiments, the proposed SSCAE model decreased after-attack accuracy to lower than 2\% (i.e., 0.56\% and 1.9\%, respectively) while keeping the perturbation percentage under 6\% (i.e., 5.3\% and 4.5\%, respectively) and the semantic similarity over 90\% (i.e., 93\% and 91\%, respectively). In the case of the BERT-Large-MNLI-Mismatched (i.e., text entailment task) experiment, the SSCAE model produced similar results to that of Table~\ref{tbl1}, where the target model is BERT.

Table \ref{tbl9} presents the experimental results of the proposed SSCAE and TEXTBUGGER adversarial attack models on 500 randomly selected testing instances from the IMDB and YELP datasets using different MLaaS platform APIs. The SSCAE model outperforms the TEXTBUGGER model in all four experiments with respect to after-attack accuracy and semantic consistency metrics, while achieving a comparable perturbation percentage. TEXTBUGGER employed a combination of word substitution and character manipulation techniques, with the character-level perturbations often changing the word meaning and resulting in a considerable reduction in semantic similarity metric. As seen, the SSCAE model produced results close to those in Table \ref{tbl1} for all evaluation metrics.     

In the majority of this study's experiments (a totally of 15 experiments), which include a variety of datasets, target models, and adversarial attack models for text classification and entailment tasks, the SSCAE model illustrated outstanding capability to generate humanly imperceptible AEs while preserving higher semantic consistency (usually $>$ 0.8) and considerably lower query number.  

\begin{figure}[h]
    \centering
    \includegraphics[width=0.49\textwidth, height= 0.45\textwidth]{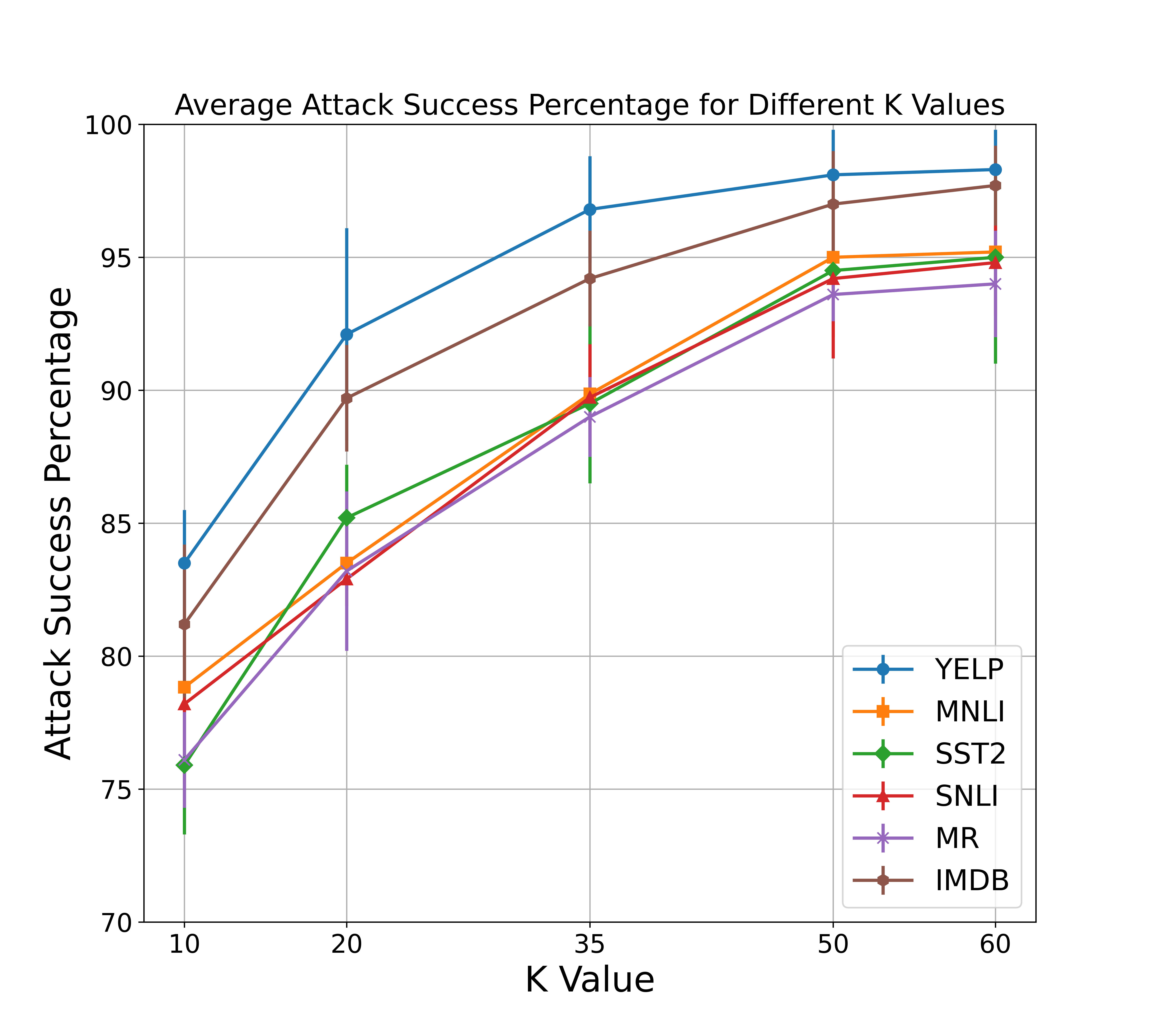}
    \caption{An ablation study of {$K$} context-aware substitutions set generated by the BERT-MLM model in step 3 of the SSCAE model}
    \label{fig2}
\end{figure}

\section{Discussion}
In this section, an ablation study is conducted to investigate some of SSCAE's major hyperparameters and strategies (i.e., different steps in Figure~\ref{fig1}). Besides, SSCAE-generated adversarial attack examples are provided. Moreover, we demonstrate our model's robustness by performing a human evaluation to verify our automatic adversarial attack experiments.

\begin{table*}[!t]
\renewcommand{\arraystretch}{1.2}
\caption{Results of an ablation study on the proposed SSCAE model using YELP (text classification task) and SNLI (text entailment task) experiments with and without semantic refinement (semantic similarity/consistency) in step 3 of the SSCAE model (Figure~\ref{fig1}) (E\#: Experiment Number; Perturb \%: Perturbation Percentage; w: With; w\/o: Without) }
\label{tbl4}
\centering
\begin{center}
\begin{tabular}{c c c c c c c}
\hline
\bfseries Dataset & \bfseries E\# & \bfseries Original Acc & \bfseries Method & \bfseries Attacked Acc & \bfseries Perturb \% & \bfseries Semantic Sim\\
\hline\hline

\multirow{2}{*}{YELP} & \multirow{2}{*}{1} & \multirow{2}{*}{95.6} & w Semantic & 1.6 (0.2) & 4.8 (0.3) & 0.92 (0.03)\\
& & & w/o Semantic & 1.2 (0.1) & 3.2 (0.2) & 0.73 (0.02) \\

\hline

\multirow{2}{*}{SNLI} & \multirow{2}{*}{5} & \multirow{2}{*}{89.4} & w Semantic & 11.5 (0.7) & 16.6 (0.9) & 0.76 (0.02)\\
& & & w/o Semantic & 7.5 (0.5) & 10.4 (0.7) & 0.60 (0.03) \\

\hline

\end{tabular}
\end{center}
\end{table*}

\begin{table*}[!t]
\renewcommand{\arraystretch}{1.2}
\caption{A comparison study between the utilization of the USE and Sentence-BERT semantic embedding models over experiments corresponding to YELP, SNLI, IMDB, and MNLI-Mismatched datasets in step 3 of the SSCAE model (Figure~\ref{fig1}). The standard deviation is included within the brackets. (E\#: Experiment Number; USE: Universal Sentence Encoder) }
\label{tbl5}
\centering
\begin{center}
\begin{tabular}{c c c c c c}
\hline
\bfseries Dataset & \bfseries E\# & \bfseries Semantic Embedding Method & \bfseries Attacked Acc & \bfseries Semantic Sim \\
\hline\hline

\multirow{2}{*}{YELP} & \multirow{2}{*}{1} & USE & 1.6 (0.2) & 0.92 (0.04) \\
& & Sentence-BERT & 2.8 (0.3) & 0.92 (0.03) \\

\hline

\multirow{2}{*}{IMDB} & \multirow{2}{*}{2} & USE & 4.2 (0.4) & 0.86 (0.03) \\
& & Sentence-BERT & 8.7 (0.7) & 0.84 (0.04) \\

\hline

\multirow{2}{*}{SNLI} & \multirow{2}{*}{5} & USE & 11.5 (1.1) & 0.76 (0.03) \\
& & Sentence-BERT & 13.6 (1.3) & 0.74 (0.02) \\

\hline

\multirow{2}{*}{MNLI-Mismatched} & \multirow{2}{*}{7} & USE & 9.6 (0.8) & 0.82 (0.04) \\
& & Sentence-BERT & 12.5 (0.9) & 0.81 (0.03) \\

\hline

\end{tabular}
\end{center}
\end{table*}

\subsection{A Comparison Study of {$K$} in Context-Aware Substitutions}
Figure~\ref{fig2} presents an ablation study of {$K$} context-aware substitutions of step 3 in Figure ~\ref{fig1}. The horizontal axis shows five possible {$K$} values of 10, 20, 35, 50, and 60, and the vertical axis indicates the Attack Success Percentage (ASP) metric in six experiments (experiments 1-5, 7) corresponding to YELP, IMDB, SST2, MR, SNLI, and MLNI-Mismatched datasets. ASP metric is computed using the AEs generated by only those testing input samples that were correctly classified by the target model for a fair comparative study. Generally, a larger {$K$} means a larger number of substitutions for an important word; it increases the chance of producing AEs (with a lower perturbation percentage) that fool the model (larger ASP) despite utilizing linguistic refinements. However, in our experiments, starting from {$K=50$}, the ASP improvement (semantic consistency reduction) rate decreases (increases). At {$K \geq 60$}, the ASP improvement rate is insignificant, while the semantic consistency reduction rate is considerable. As such, for the proposed SSCAE model, a {$K=60$} was found to be near-optimum in all experiments with reasonable semantic consistency.

\begin{table*}[!t]
\renewcommand{\arraystretch}{1.2}
\caption{A comparative study of the Average\_threshold, Median\_threshold, TopN\_threshold, and Top\_maxes\_distance heuristics to compute the dynamic threshold in step 3 of the SSCAE model (Figure ~\ref{fig1}). The standard deviation is included within the brackets. (E\#: Experiment Number; Perturb \%: Perturbation Percentage) }
\label{tbl6}
\centering
\begin{center}
\begin{tabular}{c c c c c c}
\hline
\bfseries Dataset & \bfseries E\# & \bfseries Method & \bfseries Attacked Acc & \bfseries Perturb \% & \bfseries Semantic Sim \\
\hline\hline

\multirow{5}{*}{YELP} & \multirow{5}{*}{1} &
Constant Threshold & 2.2 (0.3) & 4.8 (0.4) & 0.87 (0.02) \\
& & Average\_threshold & 3.0 (0.4) & 6.4 (0.5) & 0.90 (0.03) \\
& & Median\_threshold & 2.8 (0.4) & 6.5 (0.5) & 0.90 (0.04) \\
& & TopN\_threshold & 2.0 (0.3) & 4.7 (0.4) & 0.92 (0.03) \\
& & Top\_maxes\_distance & 1.6 (0.2) & 4.8 (0.3) & 0.92 (0.03) \\

\hline

\multirow{5}{*}{MNLI-Mismatched} & \multirow{5}{*}{7} &
Constant Threshold & 11.2 (0.8) & 13.7 (1.1) & 0.78 (0.04) \\
& & Average\_threshold & 12.8 (0.9) & 17.4 (1.4) & 0.79 (0.02) \\
& & Median\_threshold & 12.6 (0.9) & 17.5 (1.5) & 0.81 (0.03) \\
& & TopN\_threshold & 10.9 (0.8) & 13.9 (1.0) & 0.81 (0.05) \\
& & Top\_maxes\_distance & 9.6 (0.7) & 13.9 (0.9) & 0.82 (0.03) \\

\hline

\end{tabular}
\end{center}
\end{table*}

\subsection{Importance of Semantic Refinement}
Table~\ref{tbl4} presents the results of a sensitivity analysis on the SSCAE model using YELP (as an example of the text classification task) and SNLI (as an example of the text entailment task) datasets (experiments 1 and 5) with and without semantic refinement step (semantic similarity in step 3, Figure~\ref{fig1}). By removing semantic refinement, after-attack accuracy, perturbation percentage, and semantic consistency are respectively dropped from 1.6\% to 1.2\% (i.e., improved), 4.8\% to 3.2\% (i.e., improved), and 0.92 to 0.73 (i.e., worsened). Although after-attack accuracy and perturbation percentage metrics improved slightly, semantic consistency was worsened, indicating that, on average, the AEs lost their original meaning. Hence, the SSCAE-generated AEs would be humanly detectable. As such, semantic refinement plays an essential role in generating high-quality and imperceptible AEs in the proposed SSCAE model. 

In step 3 of the SSCAE model (Figure~\ref{fig1}), USE assigns a semantic similarity score to each candidate substitution. Table~\ref{tbl5} compares the utilization of USE with another possible semantic embedding model, Sentence-BERT \cite{reimers2019sentence}, over experiments corresponding to YELP, IMDB, SNLI, and MNLI-Mismatched datasets. These results indicate that USE outperforms Sentence-BERT in not only achieving a lower after-attack accuracy but a slightly higher semantic consistency. Therefore, USE was selected as the primary semantic embedding model in the proposed SSCAE model.

\begin{table*}[!t]
\renewcommand{\arraystretch}{1.2}
\caption{The efficacy of the minor hyperparameter \(\lambda\) within the Top\_maxes\_distance technique. (E\#: Experiment Number; Perturb \%: Perturbation Percentage) }
\label{tbl16}
\centering
\begin{center}
\begin{tabular}{c c c c c c}
\hline
\bfseries Dataset & \bfseries E\# & \bfseries  \(\lambda\) & \bfseries Attacked Acc & \bfseries Perturb \% & \bfseries Semantic Sim \\
\hline\hline

\multirow{4}{*}{YELP} & \multirow{4}{*}{1} &
0.5 & 1.8 (0.1) & 4.8 (0.4) & 0.90 (0.04) \\
& & 1 & 1.6 (0.2) & 4.8 (0.5) & 0.92 (0.03) \\
& & 5 & 1.9 (0.2) & 4.9 (0.3) & 0.89 (0.02) \\
& & 10 & 2.1 (0.2) & 4.9 (0.5) & 0.88 (0.02) \\

\hline

\end{tabular}
\end{center}
\end{table*}

\subsection{Dynamic Threshold Investigations}
Table~\ref{tbl6} presents a comparative study of the constant threshold and the four proposed heuristics (dynamic thresholds) for semantic and syntactic refinements in step 3 of the SSCAE model (Figure~\ref{fig1}). Average\_threshold and Median\_threshold obtained proportional results in all after-attack accuracy, perturbation percentage, and semantic consistency metrics, perhaps, because they both use similar mathematical approaches for the refinement task. On average, TopN\_threshold and Top\_maxes\_distance produced better results than Average\_threshold and Median\_threshold. However, after-attack accuracy results in Top\_maxes\_distance are better than TopN\_threshold, while both produced proximate perturbation percentage and semantic consistency results. Furthermore, we conducted an investigation to determine the optimal value for the minor hyperparameter {$\lambda$} in Top\_maxes\_distance. Table~\ref{tbl16} reveals that {$\lambda$=1} yielded superior performance compared to other values examined. Compared to the dynamic thresholds, the constant threshold achieved better perturbation percentage, worst semantic consistency, and relatively moderate after-attack accuracy results. Overall, Top\_maxes\_distance is preferred for the SSCAE model due to its superior evaluation results compared to other (constant/dynamic) threshold techniques.

\subsection{The Influence of Neighboring Words}
\begin{figure}[h]
    \centering
    \includegraphics[width=0.48\textwidth, height= 0.4\textwidth]{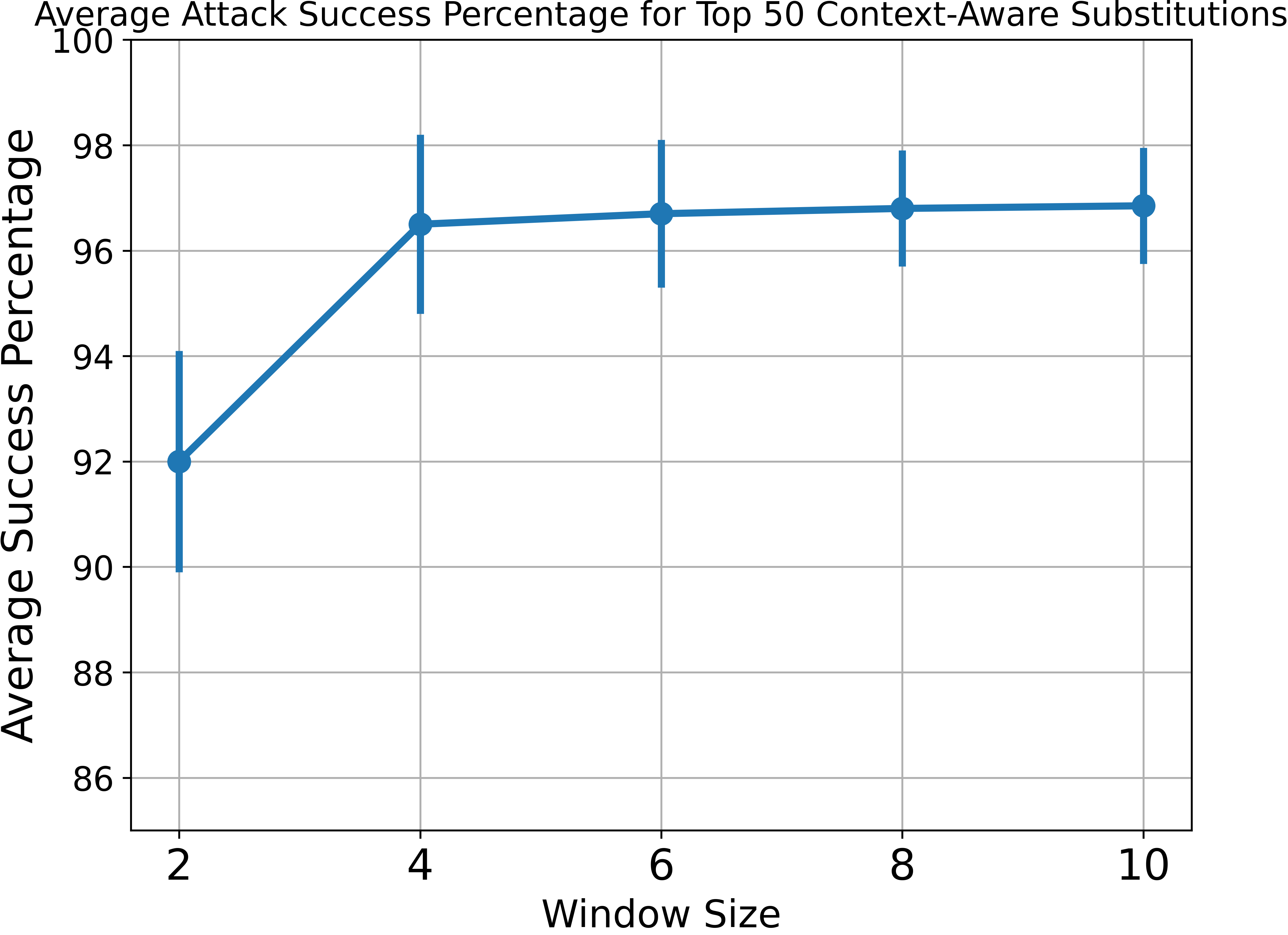}
    \caption{An ablation study of the influence of different window sizes on the retrieval of distinct sets of the top $K=50$ context-aware substitutions.}
    \label{fig5}
\end{figure}

Figure~\ref{fig5} presents an ablation study of the influence of different window sizes on the retrieval of distinct sets of the top $K=50$ context-aware substitutions. A larger window size ($|W|$) results in a greater number of substitutions, thereby increasing the likelihood of generating worst-case AEs that deceive the target model. Although there is a significant improvement in Average Success Percentage (ASP) from $|W|=2$ to $|W|=4$, starting from $|W|=4$, the rate of improvement in ASP begins to diminish. At $|W|\geq4$, the rate of ASP improvement becomes insignificant, while the search space required to find efficient adversarial samples significantly increases. Consequently, we have opted to use $|W|=4$ for refining substitutions in our experiments.

\subsection{Local Greedy Search}

Theoretically, there exists a wide range of values that can be used to set the parameters {$\mathrm{M}$} and {$\mathrm{N}$} in the local greedy search method. However, due to the significant computational cost associated with evaluating different combinations of substitutions, denoted as {$\mathrm{N}^\mathrm{M}$}, it is necessary to limit the range of these hyperparameters to smaller values. Based on our experimental observations, a practical solution in terms of computational efficiency is to set both {$\mathrm{M}$} and {$\mathrm{N}$} within the range of 1 to 4 (i.e., [1, 4]). This approach effectively limits the total number of combinations to a maximum of 256, thereby achieving a desirable trade-off between computational feasibility and comprehensive evaluation.   

\begin{figure}[t]
    \centering
    \includegraphics[width=\columnwidth]{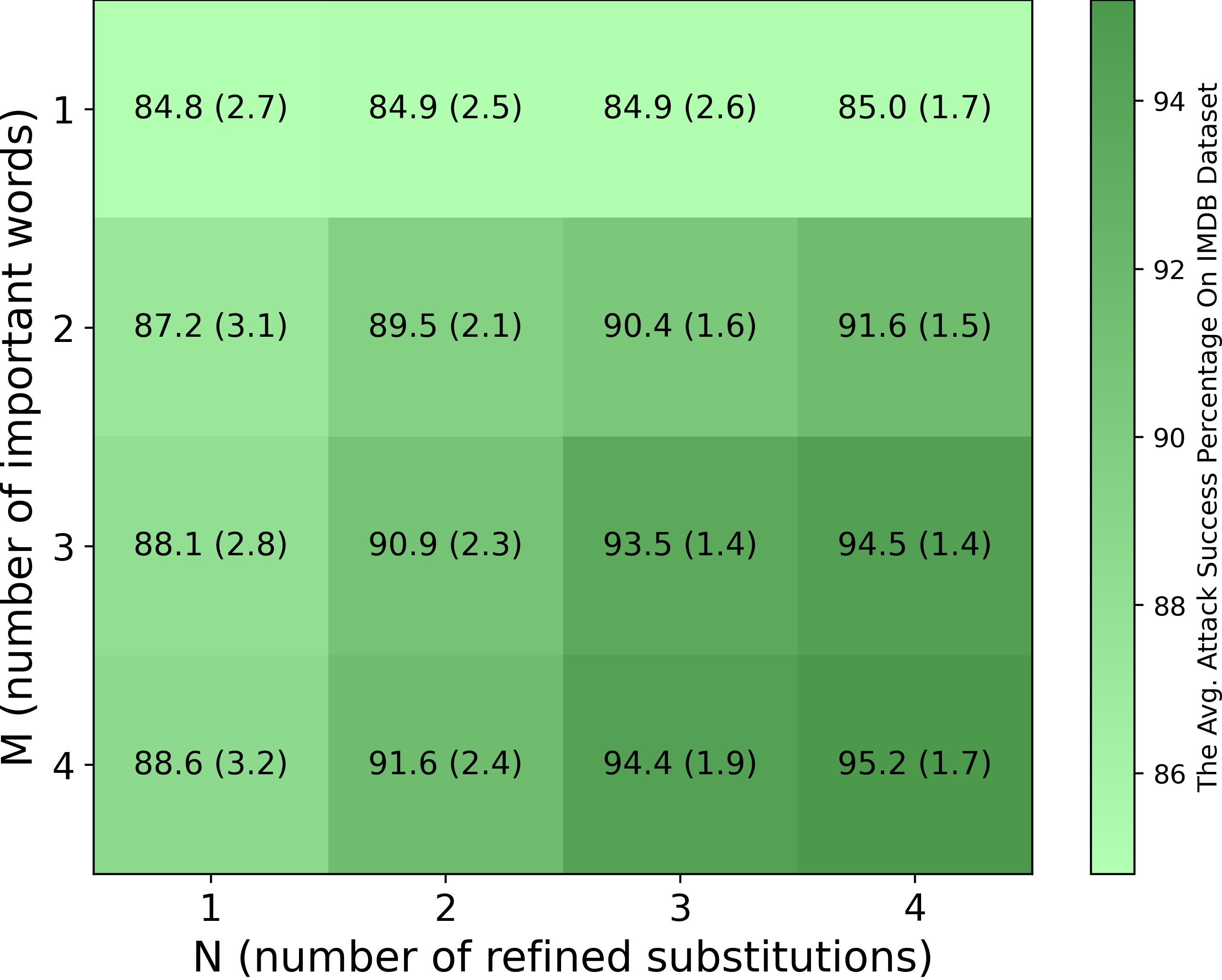}
    \caption{The impact of the number of important words and candidate substitutions on the average attack success percentage of the IMDB dataset.}
    \label{fig3}
\end{figure}

Figure {~\ref{fig3}} shows how different numbers of important words, {$\mathrm{M}$}, and refined substitutions, {$\mathrm{N}$}, can affect the average ASP benchmark. When {$\mathrm{M}$} = 1, the average ASP is around 85\% for different {$\mathrm{N}$}s (i.e., number of substitutions). Using two important words, the average ASP is increased to 91.6\% for {$\mathrm{N}$} = 4 substitutions, a considerable improvement. For a higher number of important words (e.g., {$\mathrm{M}$} = 3 or  {$\mathrm{M}$} = 4), the average ASP shows a similar pattern of improvement with different {$\mathrm{N}$}s. A larger {$\mathrm{N}$} would give a better chance of discovering AEs with higher ASPs. However, a larger {$\mathrm{N}$} exponentially increases the search space necessary for finding efficient adversarial samples. We adopt {$\mathrm{N}$} = 4 for refined substitutions and {$\mathrm{M}$} = 3 for important words in our experiments.

\begin{table*}[!t]
\renewcommand{\arraystretch}{1.2}
\caption{Examples of original and adversarial sentences from experiments corresponding to YELP and MNLI datasets (E\#: Experiment Number;  Pair \#:  Pair Number; P: Positive; N: Negative; E: Entailment; NE: Neutral) }
\label{tbl7}
\centering
\begin{center}
\begin{tabular}{ c c c c p{9cm}}
\hline

\bfseries Dataset & \bfseries E\# & \bfseries Pair \# & \bfseries Data Type (Prediction) & \bfseries Sentence \\
\hline\hline

\vspace{0.1cm}

\multirow{9}{*}{YELP} & \multirow{9}{*}{1} & \multirow{3.5}{*}{1} & Input Sample  (P) & Yes! Awesome soy cap, scone, and atmosphere. \textbf{Nice} place to hang out \& read, and free WiFi with no login procedure.\\
\vspace{0.2cm}
& & & Adversarial Example  (N) & Yes! Awesome soy cap, scone, and atmosphere. \textbf{Fantastic} place to hang out \& read, and free WiFi with no login procedure. \\

\cline{3-5}
\vspace{0.1cm}
& & \multirow{3}{*}{2} & Input Sample (N) & Refused to take my cat, which had passed away, for cremation cause I had not been to the \textbf{clinic} previously... \\
\vspace{0.2cm}
& & & Adversarial Example (P) & Refused to take my cat, which had passed away, for cremation cause I had not been to the \textbf{hospital} previously... \\

\hline
\vspace{0.1cm}
\multirow{11.5}{*}{MNLI-Mismatched} & \multirow{11.5}{*}{7} & \multirow{5.5}{*}{1} & Hypothesis & Poirot was disappointed with me \\
\vspace{0.1cm}
& & & Input Sample  (NE) & Still, it would be interesting to know. 109 Poirot looked at me very \textbf{earnestly}, and again shook his head \\
\vspace{0.2cm}
& & & Adversarial Example  (E) & Still, it would be interesting to know. 109 Poirot looked at me very \textbf{carefully}, and again shook his head \\

\cline{3-5}
\vspace{0.1cm}
& & \multirow{15}{*}{2} & Hypothesis & Talking about privacy is a complicated topic, there are a couple different ways of talking about it, for example privacy is something that disturbs your private state... \\
\vspace{0.1cm}
& & & Input Sample  (E) & Well the first thing for me is i wonder i see a couple of different ways of talking about what privacy is um if privacy is something that disturbs your private state i mean an invasion of privacy is something that disturbs your private state that’s one thing and if privacy is something that comes into your private state and extracts information from it in other words finds something out about you that’s another and the first kind of invasion of the first type of privacy seems invaded to me in very much everyday in this country but in the second type at least overtly uh where someone comes in and uh finds out information about you that should be private uh does not seem uh um obviously \textbf{everyday} \\
\vspace{0.2cm}
& & & Adversarial Example  (NE) & Well the first thing for me is i wonder i see a couple of different ways of talking about what privacy is um if privacy is something that disturbs your private state i mean an invasion of privacy is something that disturbs your private state that’s one thing and if privacy is something that comes into your private state and extracts information from it in other words finds something out about you that’s another and the first kind of invasion of the first type of privacy seems invaded to me in very much everyday in this country but in the second type at least overtly uh where someone comes in and uh finds out information about you that should be private uh does not seem uh um obviously \textbf{routine} \\

\hline

\end{tabular}
\end{center}
\end{table*}

\subsection{Examples of Adversarial Texts}
Table{~\ref{tbl7}} presents four pairs of original input samples and corresponding SSCAE-generated adversarial attack examples from experiments corresponding to YELP (two pairs) and MNLI-Mismatch (two pairs) datasets. In YELP, the first (second) pair, the adjective ``Nice'' (noun ``clinic'') in the input sample, is recognized as an important word and replaced with ``Fantastic'' (``hospital'') to generate an adversarial attack example that fools the BERT model. Although these two adjectives (nouns) are not necessarily synonyms despite arguable similarities, the general meaning of the original sample is remarkably preserved. Besides, the AE is intact grammatically and syntactically. These promising results are achieved thanks to step 3 in the SSCAE model (Figure{~\ref{fig1}}), where linguistic filters significantly improved the quality of the generated AEs in terms of imperceptibility and fluency. The same lexical analysis can be used for the generated examples related to the MNLI-Mismatch dataset.

\begin{table*}[!t]
\renewcommand{\arraystretch}{1.2}
\caption{Average results of adversarial attacks on the normal and adversarial fine-tuning of the BERT-Large model using two different attack methods: TextFooler and the proposed SSCAE. The standard deviation is included within the brackets. (E\#: Experiment Number;  Perturb \%: Perturbation Percentage)}

\label{tbl13}
\centering
\begin{center}
\begin{tabular}{c c c c c c c}
\hline
\bfseries Dataset & \bfseries E\# & \bfseries Train Method & \bfseries Original Acc & \bfseries Adversarial Attack & \bfseries Attacked Acc & \bfseries Perturb \% \\
\hline\hline

\multirow{3}{*}{MR} &
\multirow{3}{*}{4} &
Normal Fine-tune & 85.3 & SSCAE (ours) & 15.6 (1.0) & 17.4 (0.9) \\
 & & Adversarial Fine-tune & 84.8 & TextFooler & 38.2 (1.7) & 24.6 (1.3)\\
 & & Adversarial Fine-tune & 84.8 & SSCAE (ours) & {32.4 (1.5)} & 21.6 (1.1) \\
\hline

\multirow{3}{*}{SNLI} &
\multirow{3}{*}{5} &
Normal Fine-tune & 89.4 & SSCAE (ours) & 11.5 (0.6) & 16.6 (0.8) \\
 & & Adversarial Fine-tune & 88.9 & TextFooler & 26.1 (1.4) & 23.9 (1.4)\\
 & & Adversarial Fine-tune & 88.9 & SSCAE (ours) & {21.2 (1.2)} & 20.5 (1.0) \\

\hline

\end{tabular}
\end{center}
\end{table*}

\subsection{Adversarial Training}
To implement an effective defense mechanism, we employ the adversarial training technique to enhance the robustness of the target models against various adversarial attacks, particularly focusing on the proposed SSCAE model. Additionally, we perform fine-tuning on a target model (in this case, the BERT-Large model) by providing it with both the original data and the adversarial examples generated by the SSCAE model. We have collected adversarial examples from the MR and SNLI training sets, which successfully deceived BERT-Large, and incorporated them into the original training set. Subsequently, the augmented data is utilized for fine-tuning the target model. Our objective is to evaluate the resilience of this adversarially fine-tuned model against two distinct adversarial attack models: TextFooler and SSCAE. 

In Table~\ref{tbl13}, the after-attack accuracy and perturbation ratio are observed to increase after adversarial fine-tuning for both the TextFooler and SSCAE attack methods. Furthermore, the original accuracy of the adversarially-trained model remains comparable to the model trained on clean datasets. These findings indicate that adversarial training using a high-quality set of adversarial examples renders the attack more challenging, highlighting the increased difficulty of compromising the model's robustness. Notably, the target model exhibits greater resilience against TextFooler compared to the proposed SSCAE model, suggesting that the robustness of a target model against future attacks can be enhanced by training it with the efficient adversarial examples generated by the SSCAE model.

\begin{table*}[!t]
\renewcommand{\arraystretch}{1.2}
\caption{Human Evaluation Task (E\#: Experiment Number) }
\label{tbl3}
\centering
\begin{center}
\begin{tabular}{c c c c c c}
\hline
\bfseries Dataset & \bfseries Adversarial Attack  & \bfseries Data Type & \bfseries Human Accuracy & \bfseries Meaningfulness & \bfseries Grammar Correctness \\
\hline\hline

\multirow{5}{*}{YELP} & - & Original & 92.5 & 4.2 & 4.0 \\
& {TextFooler} & Adversarial & 83.4 & 3.1 & 3.2 \\
& {BERT-Attack} & Adversarial & 84.7 & 3.8 & 3.6 \\
& {BAE} & Adversarial & 84.8 & 3.7 & 3.5 \\
& {SSCAE (ours)} & Adversarial & 86.0 & 4.0 & 3.8 \\

\hline

\multirow{5}{*}{MNLI-Mismatched} & - & Original & 91.2 & 3.9 & 4.1 \\
& {TextFooler} & Adversarial & 74.2 & 3.0 & 3.1 \\
& {BERT-Attack} & Adversarial & 75.6 & 3.5 & 3.6 \\
& {BAE} & Adversarial & 75.3 & 3.5 & 3.5 \\
& {SSCAE (ours)} & Adversarial & 77.4 & 3.7 & 3.7 \\

\hline

\end{tabular}
\end{center}
\end{table*}

\subsection{Human Evaluation}
Human evaluation is a useful study to weigh the AEs correctness in terms of imperceptibility, language fluency, and semantic consistency. However, only a few studies \cite{li2020bert} performed comprehensive human evaluations. Table~\ref{tbl3} presents a human assessment of the quality and fluency of AEs generated by the proposed SSCAE, TextFooler, BERT-Attack, and BAE adversarial attack models on YELP, as an example of the text classification task, and MLNI-mismatched, as an example of the text alignment task with BERT as the target model (i.e., experiments 1 and 7). In each experiment, three PhD candidates from Georgia State University, Department of Applied Linguistics, were chosen for their robust scientific backgrounds. These candidates, who maintain independence from the authors of this paper, provided 100 randomly selected input samples (denoted as ``Original'' in Table~\ref{tbl3}) and their corresponding adversarial samples generated by different adversarial attack methods (denoted as ``Adversarial'' in Table~\ref{tbl3}). The Original and Adversarial samples were shuffled. Whether Original or Adversarial, for a particular sample, if the majority of students correctly estimate the class of a sample, it is counted as one correct human estimation. While the SSCAE model outperforms other adversarial attack methods, there is only a small gap (6.5\%) between the human estimation of the Original samples and the SSCAE-generated Adversarial samples in the YELP experiment, indicating SSCAE's capability to generate high-quality and imperceptible adversarial samples. Even though the SSCAE-generated adversarial samples achieve better human accuracy than other methods, this gap is larger in MNLI-Mismatched because human-crafted hypothesis and premise sentences share a considerable amount of the same words, and applying perturbations on these words would negatively affect the human assessment to make the correct prediction.

Furthermore, each student is asked to provide two Likert scores (1 to 5) for each sample, whether Original or Adversarial. The first score is about how meaningful (1 to be meaningless and 5 to be meaningful) the sample is, and the second score represents the extent of the sample's grammar correctness (1 to be incorrect and 5 to be correct). The SSCAE-generated adversarial samples achieve higher (better) scores in both meaningfulness and grammatical correctness metrics compared to the adversarial samples generated by other state-of-the-art adversarial methods. In terms of meaningfulness (grammatical correctness), there is only a small gap of 0.2 (0.2 and 0.4) between the Original and Adversarial scores in both YELP and MNLI-Mismatched experiments; the SSCAE-generated adversarial samples are semantically and grammatically within the same distribution as the Original samples.

\section{Conclusion}
In this work, we introduced SSCAE, a practical black-box AE generator for developing context-wise AEs while preserving essential linguistic features (semantic, syntactic, and grammatical). The SSCAE utilizes the BERT MLM model to generate potential substitutions for each important word. Besides, it employs three refinement techniques to maintain the linguistic properties of final perturbations. We introduced (1) a thresholding technique, dubbed dynamic threshold, to capture more effective perturbations and (2) a word permutation-substitution technique, dubbed local greedy search, to generate high-quality adversarial examples. The SSCAE model was compared with three state-of-the-art black-box methods that are TextFooler, BERT-Attack, and BAE. The SSCAE model outperformed these models, with a considerable margin over different popular and challenging text classification and text entailment tasks, with regard to evaluation benchmarks such as after-attack accuracy, average query number, and average semantic consistency. In terms of average perturbation percentage, SSCAE outperformed four out of seven NLP tasks. Besides, the human evaluation showed the effectiveness of our proposed model in generating high-quality and imperceptible perturbations. The SSCAE model can be flexibly extended with different advanced components due to its fully modular characteristic: it is not limited to USE and GPT-2. One can replace the SSCAE's USE and GPT-2 components with other (perhaps, more efficient) language models to extract more effective perturbations in terms of semantic and syntactic features. Moreover, BERT MLM can be replaced with Transformer-based models such as Roberta, Albert, and BART \cite{lewis2020bart} to, perhaps, obtain high-quality contextualized substitutions. Finally, the dynamic threshold technique and local greedy search can each be substituted with, perhaps, complicated but still efficient techniques, such as mixed-integer optimization, to achieve better performance. However, the SSCAE model is complex and requires relatively high computational resources. Developing a parallel computing interface to boost the computational efficiency of the proposed model would be a future direction. Besides, implementing practical operations in similar studies, such as insertion and deletion, within the SSCAE model remains an open question ripe for further investigation.

% or


\appendices
\section{Dataset Description}
\label{sec:appendix1}
\textbf{YELP} (business) is a document-level dataset with 560,000 training and 38,000 testing highly polar samples where negative and positive classes are 1- and 2-star  and 4- and 5-star reviews, respectively.

\textbf{IMDb} Review (movie) is a document-level dataset with 25,000 training and 25,000 testing highly polar samples where negative and positive classes are review scores $\leq$4 and $\geq$7 out of 10, respectively.

\textbf{RTMR} (movie) is a sentence-level dataset based on sentiment polarity with 8530 training and 1066 testing highly polar samples where negative and positive classes are assigned based on the calibration among different critics. 

\textbf{SST2} (movie) is a sentence-level dataset based on sentiment polarity with 8544 training and 2210 testing highly polar samples where any multi-level negative and positive reviews are categorized as negative and positive reviews (neutral reviews are excluded). 

\textbf{SNLI (MNLI)} is a three-class dataset of 550,152 (392,702) training and 10,000 (19,643) testing human-written sentence pairs in English. Every three pairs of SNLI (MLNI) are created using a different image caption from the Flicker30K dataset \cite{young2014image} (ten sources of text), called a premise sentence \cite{bowman2015large}. The premise sentence is the first sentence in each of three pairs. The second sentence (called a hypothesis sentence) \cite{bowman2015large} of the first, second, and third pair is generated to be in entailment (category 1), contradiction (category 2), and neutral (category 3) with the premise sentence, respectively. In contrast with SNLI, where premise sentences are from a relatively homogeneous image caption dataset, MNLI covers broader text styles \cite{williams2017broad}. MLNI testing sample pairs are divided into two general categories, “Matched” and “Mismatched;” the MNLI-Matched testing pairs, in contrast to MNLI-Mismatched, share similar context and resemblance as the training pairs.

\ifCLASSOPTIONcompsoc
  \section*{Acknowledgments}
\else
  \section*{Acknowledgment}
\fi

This work was supported in part by the National Science Foundation under Grants 2413654, 2054968 and in part by the Microsoft Faculty Fellowship Program.

\ifCLASSOPTIONcaptionsoff
  \newpage
\fi



%\bibliographystyle{IEEEtran}
\bibliographystyle{IEEEtran}
\bibliography{Main_Content}

\begin{thebibliography}{10}
\providecommand{\url}[1]{#1}
\csname url@samestyle\endcsname
\providecommand{\newblock}{\relax}
\providecommand{\bibinfo}[2]{#2}
\providecommand{\BIBentrySTDinterwordspacing}{\spaceskip=0pt\relax}
\providecommand{\BIBentryALTinterwordstretchfactor}{4}
\providecommand{\BIBentryALTinterwordspacing}{\spaceskip=\fontdimen2\font plus
\BIBentryALTinterwordstretchfactor\fontdimen3\font minus \fontdimen4\font\relax}
\providecommand{\BIBforeignlanguage}[2]{{%
\expandafter\ifx\csname l@#1\endcsname\relax
\typeout{** WARNING: IEEEtran.bst: No hyphenation pattern has been}%
\typeout{** loaded for the language `#1'. Using the pattern for}%
\typeout{** the default language instead.}%
\else
\language=\csname l@#1\endcsname
\fi
#2}}
\providecommand{\BIBdecl}{\relax}
\BIBdecl

\bibitem{goodfellow2014explaining}
\BIBentryALTinterwordspacing
I.~J. Goodfellow, J.~Shlens, and C.~Szegedy, ``Explaining and harnessing adversarial examples,'' in \emph{3rd International Conference on Learning Representations, {ICLR} 2015, San Diego, CA, USA, May 7-9, 2015, Conference Track Proceedings}, Y.~Bengio and Y.~LeCun, Eds., 2015. [Online]. Available: \url{http://arxiv.org/abs/1412.6572}
\BIBentrySTDinterwordspacing

\bibitem{kurakin2016adversarial}
A.~Kurakin, I.~J. Goodfellow, and S.~Bengio, ``Adversarial examples in the physical world,'' in \emph{Artificial intelligence safety and security}.\hskip 1em plus 0.5em minus 0.4em\relax Chapman and Hall/CRC, 2018, pp. 99--112.

\bibitem{zhang2020adversarial}
W.~E. Zhang, Q.~Z. Sheng, A.~Alhazmi, and C.~Li, ``Adversarial attacks on deep-learning models in natural language processing: A survey,'' \emph{ACM Transactions on Intelligent Systems and Technology (TIST)}, vol.~11, no.~3, pp. 1--41, 2020.

\bibitem{shafahi2020universal}
A.~Shafahi, M.~Najibi, Z.~Xu, J.~Dickerson, L.~S. Davis, and T.~Goldstein, ``Universal adversarial training,'' in \emph{Proceedings of the AAAI Conference on Artificial Intelligence}, vol.~34, no.~04, 2020, pp. 5636--5643.

\bibitem{xu2020adversarial}
H.~Xu, Y.~Ma, H.-C. Liu, D.~Deb, H.~Liu, J.-L. Tang, and A.~K. Jain, ``Adversarial attacks and defenses in images, graphs and text: A review,'' \emph{International Journal of Automation and Computing}, vol.~17, no.~2, pp. 151--178, 2020.

\bibitem{wang2021convergence}
\BIBentryALTinterwordspacing
Y.~Wang, X.~Ma, J.~Bailey, J.~Yi, B.~Zhou, and Q.~Gu, ``On the convergence and robustness of adversarial training,'' in \emph{Proceedings of the 36th International Conference on Machine Learning}, ser. Proceedings of Machine Learning Research, K.~Chaudhuri and R.~Salakhutdinov, Eds., vol.~97.\hskip 1em plus 0.5em minus 0.4em\relax PMLR, 09--15 Jun 2019, pp. 6586--6595. [Online]. Available: \url{https://proceedings.mlr.press/v97/wang19i.html}
\BIBentrySTDinterwordspacing

\bibitem{papernot2017practical}
N.~Papernot, P.~McDaniel, I.~Goodfellow, S.~Jha, Z.~B. Celik, and A.~Swami, ``Practical black-box attacks against machine learning,'' in \emph{Proceedings of the 2017 ACM on Asia conference on computer and communications security}, 2017, pp. 506--519.

\bibitem{chakraborty2018adversarial}
\BIBentryALTinterwordspacing
A.~Chakraborty, M.~Alam, V.~Dey, A.~Chattopadhyay, and D.~Mukhopadhyay, ``A survey on adversarial attacks and defences,'' \emph{{CAAI} Transactions on Intelligence Technology}, vol.~6, no.~1, pp. 25--45, Mar. 2021. [Online]. Available: \url{https://doi.org/10.1049/cit2.12028}
\BIBentrySTDinterwordspacing

\bibitem{jin2020bert}
D.~Jin, Z.~Jin, J.~T. Zhou, and P.~Szolovits, ``Is bert really robust? a strong baseline for natural language attack on text classification and entailment,'' in \emph{Proceedings of the AAAI conference on artificial intelligence}, vol.~34, no.~05, 2020, pp. 8018--8025.

\bibitem{li2020bert}
\BIBentryALTinterwordspacing
L.~Li, R.~Ma, Q.~Guo, X.~Xue, and X.~Qiu, ``{BERT}-{ATTACK}: Adversarial attack against {BERT} using {BERT},'' in \emph{Proceedings of the 2020 Conference on Empirical Methods in Natural Language Processing ({EMNLP})}.\hskip 1em plus 0.5em minus 0.4em\relax Association for Computational Linguistics, 2020. [Online]. Available: \url{https://doi.org/10.18653/v1/2020.emnlp-main.500}
\BIBentrySTDinterwordspacing

\bibitem{song2020universal}
\BIBentryALTinterwordspacing
L.~Song, X.~Yu, H.-T. Peng, and K.~Narasimhan, ``Universal adversarial attacks with natural triggers for text classification,'' in \emph{Proceedings of the 2021 Conference of the North American Chapter of the Association for Computational Linguistics: Human Language Technologies}.\hskip 1em plus 0.5em minus 0.4em\relax Online: Association for Computational Linguistics, Jun. 2021, pp. 3724--3733. [Online]. Available: \url{https://aclanthology.org/2021.naacl-main.291}
\BIBentrySTDinterwordspacing

\bibitem{garg2020bae}
\BIBentryALTinterwordspacing
S.~Garg and G.~Ramakrishnan, ``{BAE}: {BERT}-based adversarial examples for text classification,'' in \emph{Proceedings of the 2020 Conference on Empirical Methods in Natural Language Processing (EMNLP)}.\hskip 1em plus 0.5em minus 0.4em\relax Online: Association for Computational Linguistics, Nov. 2020, pp. 6174--6181. [Online]. Available: \url{https://aclanthology.org/2020.emnlp-main.498}
\BIBentrySTDinterwordspacing

\bibitem{devlin2018bert}
J.~Devlin, M.-W. Chang, K.~Lee, and K.~Toutanova, ``Bert: Pre-training of deep bidirectional transformers for language understanding,'' in \emph{Proceedings of the 2019 Conference of the North American Chapter of the Association for Computational Linguistics: Human Language Technologies (NAACL-HLT)}, 2019, pp. 4171--4186.

\bibitem{cer2018universal}
D.~Cer, Y.~Yang, S.-y. Kong, N.~Hua, N.~Limtiaco, R.~S. John, N.~Constant, M.~Guajardo-Cespedes, S.~Yuan, C.~Tar \emph{et~al.}, ``Universal sentence encoder for english,'' in \emph{Proceedings of the 2018 Conference on Empirical Methods in Natural Language Processing: System Demonstrations}, 2018, pp. 169--174.

\bibitem{radford2019language}
A.~Radford, J.~Wu, R.~Child, D.~Luan, D.~Amodei, I.~Sutskever \emph{et~al.}, ``Language models are unsupervised multitask learners,'' \emph{OpenAI blog}, vol.~1, no.~8, p.~9, 2019.

\bibitem{li2018textbugger}
\BIBentryALTinterwordspacing
J.~Li, S.~Ji, T.~Du, B.~Li, and T.~Wang, ``{TextBugger}: Generating adversarial text against real-world applications,'' in \emph{Proceedings 2019 Network and Distributed System Security Symposium}.\hskip 1em plus 0.5em minus 0.4em\relax Internet Society, 2019. [Online]. Available: \url{https://doi.org/10.14722/ndss.2019.23138}
\BIBentrySTDinterwordspacing

\bibitem{jia2017adversarial}
\BIBentryALTinterwordspacing
R.~Jia and P.~Liang, ``Adversarial examples for evaluating reading comprehension systems,'' in \emph{Proceedings of the 2017 Conference on Empirical Methods in Natural Language Processing}.\hskip 1em plus 0.5em minus 0.4em\relax Copenhagen, Denmark: Association for Computational Linguistics, Sep. 2017, pp. 2021--2031. [Online]. Available: \url{https://aclanthology.org/D17-1215}
\BIBentrySTDinterwordspacing

\bibitem{maheshwary2021generating}
R.~Maheshwary, S.~Maheshwary, and V.~Pudi, ``Generating natural language attacks in a hard label black box setting,'' in \emph{Proceedings of the 35th AAAI Conference on Artificial Intelligence}, 2021.

\bibitem{liu2023aliasing}
J.~Liu, H.~Jin, G.~Xu, M.~Lin, T.~Wu, M.~Nour, F.~Alenezi, A.~Alhudhaif, and K.~Polat, ``Aliasing black box adversarial attack with joint self-attention distribution and confidence probability,'' \emph{Expert Systems with Applications}, vol. 214, p. 119110, 2023.

\bibitem{lee2022query}
D.~Lee, S.~Moon, J.~Lee, and H.~O. Song, ``Query-efficient and scalable black-box adversarial attacks on discrete sequential data via bayesian optimization,'' in \emph{International Conference on Machine Learning}.\hskip 1em plus 0.5em minus 0.4em\relax PMLR, 2022, pp. 12\,478--12\,497.

\bibitem{wu2023prada}
C.~Wu, R.~Zhang, J.~Guo, M.~De~Rijke, Y.~Fan, and X.~Cheng, ``Prada: Practical black-box adversarial attacks against neural ranking models,'' \emph{ACM Transactions on Information Systems}, vol.~41, no.~4, pp. 1--27, 2023.

\bibitem{fursov2022differentiable}
I.~Fursov, A.~Zaytsev, P.~Burnyshev, E.~Dmitrieva, N.~Klyuchnikov, A.~Kravchenko, E.~Artemova, E.~Komleva, and E.~Burnaev, ``A differentiable language model adversarial attack on text classifiers,'' \emph{IEEE Access}, vol.~10, pp. 17\,966--17\,976, 2022.

\bibitem{liu2022order}
J.~Liu, Y.~Kang, D.~Tang, K.~Song, C.~Sun, X.~Wang, W.~Lu, and X.~Liu, ``Order-disorder: Imitation adversarial attacks for black-box neural ranking models,'' in \emph{Proceedings of the 2022 ACM SIGSAC Conference on Computer and Communications Security}, 2022, pp. 2025--2039.

\bibitem{lu2023black}
S.~Lu, M.~Wang, D.~Wang, X.~Wei, S.~Xiao, Z.~Wang, N.~Han, and L.~Wang, ``Black-box attacks against log anomaly detection with adversarial examples,'' \emph{Information Sciences}, vol. 619, pp. 249--262, 2023.

\bibitem{ebrahimi2017hotflip}
\BIBentryALTinterwordspacing
J.~Ebrahimi, A.~Rao, D.~Lowd, and D.~Dou, ``{H}ot{F}lip: White-box adversarial examples for text classification,'' in \emph{Proceedings of the 56th Annual Meeting of the Association for Computational Linguistics (Volume 2: Short Papers)}.\hskip 1em plus 0.5em minus 0.4em\relax Melbourne, Australia: Association for Computational Linguistics, Jul. 2018, pp. 31--36. [Online]. Available: \url{https://aclanthology.org/P18-2006}
\BIBentrySTDinterwordspacing

\bibitem{zhao2018adversarially}
J.~Zhao, Y.~Kim, K.~Zhang, A.~Rush, and Y.~LeCun, ``Adversarially regularized autoencoders,'' in \emph{International conference on machine learning}.\hskip 1em plus 0.5em minus 0.4em\relax PMLR, 2018, pp. 5902--5911.

\bibitem{guo2021gradient}
\BIBentryALTinterwordspacing
C.~Guo, A.~Sablayrolles, H.~J{\'e}gou, and D.~Kiela, ``Gradient-based adversarial attacks against text transformers,'' in \emph{Proceedings of the 2021 Conference on Empirical Methods in Natural Language Processing}.\hskip 1em plus 0.5em minus 0.4em\relax Online and Punta Cana, Dominican Republic: Association for Computational Linguistics, Nov. 2021, pp. 5747--5757. [Online]. Available: \url{https://aclanthology.org/2021.emnlp-main.464}
\BIBentrySTDinterwordspacing

\bibitem{jang2016categorical}
\BIBentryALTinterwordspacing
E.~Jang, S.~Gu, and B.~Poole, ``Categorical reparameterization with gumbel-softmax,'' in \emph{International Conference on Learning Representations}, 2017. [Online]. Available: \url{https://openreview.net/forum?id=rkE3y85ee}
\BIBentrySTDinterwordspacing

\bibitem{zhang2019bertscore}
\BIBentryALTinterwordspacing
T.~Zhang*, V.~Kishore*, F.~Wu*, K.~Q. Weinberger, and Y.~Artzi, ``Bertscore: Evaluating text generation with bert,'' in \emph{International Conference on Learning Representations}, 2020. [Online]. Available: \url{https://openreview.net/forum?id=SkeHuCVFDr}
\BIBentrySTDinterwordspacing

\bibitem{chen2022unsupervised}
J.~Chen, C.~Gan, S.~Cheng, H.~Zhou, Y.~Xiao, and L.~Li, ``Unsupervised editing for counterfactual stories,'' in \emph{Proceedings of the AAAI Conference on Artificial Intelligence}, vol.~36, no.~10, 2022, pp. 10\,473--10\,481.

\bibitem{pang_etal_2020_towards}
\BIBentryALTinterwordspacing
B.~Pang, E.~Nijkamp, W.~Han, L.~Zhou, Y.~Liu, and K.~Tu, ``Towards holistic and automatic evaluation of open-domain dialogue generation,'' in \emph{Proceedings of the 58th Annual Meeting of the Association for Computational Linguistics}.\hskip 1em plus 0.5em minus 0.4em\relax Online: Association for Computational Linguistics, Jul. 2020, pp. 3619--3629. [Online]. Available: \url{https://aclanthology.org/2020.acl-main.333}
\BIBentrySTDinterwordspacing

\bibitem{gao2018black}
J.~Gao, J.~Lanchantin, M.~L. Soffa, and Y.~Qi, ``Black-box generation of adversarial text sequences to evade deep learning classifiers,'' in \emph{2018 IEEE Security and Privacy Workshops (SPW)}.\hskip 1em plus 0.5em minus 0.4em\relax IEEE, 2018, pp. 50--56.

\bibitem{jordan2015machine}
M.~I. Jordan and T.~M. Mitchell, ``Machine learning: Trends, perspectives, and prospects,'' \emph{Science}, vol. 349, no. 6245, pp. 255--260, 2015.

\bibitem{lecun2015deep}
Y.~LeCun, Y.~Bengio, and G.~Hinton, ``Deep learning,'' \emph{nature}, vol. 521, no. 7553, pp. 436--444, 2015.

\bibitem{mahesh2020machine}
B.~Mahesh, ``Machine learning algorithms-a review,'' \emph{International Journal of Science and Research (IJSR).[Internet]}, vol.~9, pp. 381--386, 2020.

\bibitem{liu2019roberta}
\BIBentryALTinterwordspacing
Y.~Liu, M.~Ott, N.~Goyal, J.~Du, M.~Joshi, D.~Chen, O.~Levy, M.~Lewis, L.~Zettlemoyer, and V.~Stoyanov, ``Ro{\{}bert{\}}a: A robustly optimized {\{}bert{\}} pretraining approach,'' 2020. [Online]. Available: \url{https://openreview.net/forum?id=SyxS0T4tvS}
\BIBentrySTDinterwordspacing

\bibitem{lan2019albert}
\BIBentryALTinterwordspacing
Z.~Lan, M.~Chen, S.~Goodman, K.~Gimpel, P.~Sharma, and R.~Soricut, ``Albert: A lite bert for self-supervised learning of language representations,'' in \emph{International Conference on Learning Representations}, 2020. [Online]. Available: \url{https://openreview.net/forum?id=H1eA7AEtvS}
\BIBentrySTDinterwordspacing

\bibitem{NEURIPS2019_9015}
A.~Paszke, S.~Gross, F.~Massa, A.~Lerer, J.~Bradbury, G.~Chanan, T.~Killeen, Z.~Lin, N.~Gimelshein, L.~Antiga, A.~Desmaison, A.~Kopf, E.~Yang, Z.~DeVito, M.~Raison, A.~Tejani, S.~Chilamkurthy, B.~Steiner, L.~Fang, J.~Bai, and S.~Chintala, ``Pytorch: An imperative style, high-performance deep learning library,'' in \emph{Advances in Neural Information Processing Systems 32}.\hskip 1em plus 0.5em minus 0.4em\relax Curran Associates, Inc., 2019, pp. 8024--8035.

\bibitem{spacy2}
M.~Honnibal and I.~Montani, ``{spaCy 2}: Natural language understanding with {B}loom embeddings, convolutional neural networks and incremental parsing,'' 2017, to appear.

\bibitem{rehurek2011gensim}
R.~Rehurek and P.~Sojka, ``Gensim--python framework for vector space modelling,'' \emph{NLP Centre, Faculty of Informatics, Masaryk University, Brno, Czech Republic}, vol.~3, no.~2, 2011.

\bibitem{wolf_etal_2020_transformers}
\BIBentryALTinterwordspacing
T.~Wolf, L.~Debut, V.~Sanh, J.~Chaumond, C.~Delangue, A.~Moi, P.~Cistac, T.~Rault, R.~Louf, M.~Funtowicz, J.~Davison, S.~Shleifer, P.~von Platen, C.~Ma, Y.~Jernite, J.~Plu, C.~Xu, T.~L. Scao, S.~Gugger, M.~Drame, Q.~Lhoest, and A.~M. Rush, ``Transformers: State-of-the-art natural language processing,'' in \emph{Proceedings of the 2020 Conference on Empirical Methods in Natural Language Processing: System Demonstrations}.\hskip 1em plus 0.5em minus 0.4em\relax Online: Association for Computational Linguistics, Oct. 2020, pp. 38--45. [Online]. Available: \url{https://www.aclweb.org/anthology/2020.emnlp-demos.6}
\BIBentrySTDinterwordspacing

\bibitem{morris2020textattack}
J.~Morris, E.~Lifland, J.~Y. Yoo, J.~Grigsby, D.~Jin, and Y.~Qi, ``Textattack: A framework for adversarial attacks, data augmentation, and adversarial training in nlp,'' in \emph{Proceedings of the 2020 Conference on Empirical Methods in Natural Language Processing: System Demonstrations}, 2020, pp. 119--126.

\bibitem{zhang2015character}
X.~Zhang, J.~Zhao, and Y.~LeCun, ``Character-level convolutional networks for text classification,'' \emph{Advances in neural information processing systems}, vol.~28, pp. 649--657, 2015.

\bibitem{IMDb}
\BIBentryALTinterwordspacing
A.~L. Maas, R.~E. Daly, P.~T. Pham, D.~Huang, A.~Y. Ng, and C.~Potts, ``Learning word vectors for sentiment analysis,'' in \emph{Proceedings of the 49th Annual Meeting of the Association for Computational Linguistics: Human Language Technologies}.\hskip 1em plus 0.5em minus 0.4em\relax Portland, Oregon, USA: Association for Computational Linguistics, June 2011, pp. 142--150. [Online]. Available: \url{http://www.aclweb.org/anthology/P11-1015}
\BIBentrySTDinterwordspacing

\bibitem{pang2005seeing}
\BIBentryALTinterwordspacing
B.~Pang and L.~Lee, ``Seeing stars: Exploiting class relationships for sentiment categorization with respect to rating scales,'' in \emph{Proceedings of the 43rd Annual Meeting of the Association for Computational Linguistics ({ACL}{'}05)}.\hskip 1em plus 0.5em minus 0.4em\relax Ann Arbor, Michigan: Association for Computational Linguistics, Jun. 2005, pp. 115--124. [Online]. Available: \url{https://aclanthology.org/P05-1015}
\BIBentrySTDinterwordspacing

\bibitem{socher2013recursive}
R.~Socher, A.~Perelygin, J.~Wu, J.~Chuang, C.~D. Manning, A.~Y. Ng, and C.~Potts, ``Recursive deep models for semantic compositionality over a sentiment treebank,'' in \emph{Proceedings of the 2013 conference on empirical methods in natural language processing}, 2013, pp. 1631--1642.

\bibitem{bowman2015large}
\BIBentryALTinterwordspacing
S.~R. Bowman, G.~Angeli, C.~Potts, and C.~D. Manning, ``A large annotated corpus for learning natural language inference,'' in \emph{Proceedings of the 2015 Conference on Empirical Methods in Natural Language Processing}.\hskip 1em plus 0.5em minus 0.4em\relax Lisbon, Portugal: Association for Computational Linguistics, Sep. 2015, pp. 632--642. [Online]. Available: \url{https://aclanthology.org/D15-1075}
\BIBentrySTDinterwordspacing

\bibitem{williams2017broad}
\BIBentryALTinterwordspacing
A.~Williams, N.~Nangia, and S.~Bowman, ``A broad-coverage challenge corpus for sentence understanding through inference,'' in \emph{Proceedings of the 2018 Conference of the North {A}merican Chapter of the Association for Computational Linguistics: Human Language Technologies, Volume 1 (Long Papers)}.\hskip 1em plus 0.5em minus 0.4em\relax New Orleans, Louisiana: Association for Computational Linguistics, Jun. 2018, pp. 1112--1122. [Online]. Available: \url{https://aclanthology.org/N18-1101}
\BIBentrySTDinterwordspacing

\bibitem{Google_BERT}
``Google bert,'' \url{https://abeyon.com/googles-bert-nlp/}, accessed: 2022-05-04.

\bibitem{SQuAD}
\BIBentryALTinterwordspacing
P.~Rajpurkar, J.~Zhang, K.~Lopyrev, and P.~Liang, ``{SQ}u{AD}: 100,000+ questions for machine comprehension of text,'' in \emph{Proceedings of the 2016 Conference on Empirical Methods in Natural Language Processing}.\hskip 1em plus 0.5em minus 0.4em\relax Austin, Texas: Association for Computational Linguistics, Nov. 2016, pp. 2383--2392. [Online]. Available: \url{https://aclanthology.org/D16-1264}
\BIBentrySTDinterwordspacing

\bibitem{hochreiter1997long}
S.~Hochreiter and J.~Schmidhuber, ``Long short-term memory,'' \emph{Neural computation}, vol.~9, no.~8, pp. 1735--1780, 1997.

\bibitem{chen2016enhanced}
\BIBentryALTinterwordspacing
Q.~Chen, X.~Zhu, Z.-H. Ling, S.~Wei, H.~Jiang, and D.~Inkpen, ``Enhanced {LSTM} for natural language inference,'' in \emph{Proceedings of the 55th Annual Meeting of the Association for Computational Linguistics (Volume 1: Long Papers)}.\hskip 1em plus 0.5em minus 0.4em\relax Vancouver, Canada: Association for Computational Linguistics, Jul. 2017, pp. 1657--1668. [Online]. Available: \url{https://aclanthology.org/P17-1152}
\BIBentrySTDinterwordspacing

\bibitem{reimers2019sentence}
N.~Reimers and I.~Gurevych, ``Sentence-bert: Sentence embeddings using siamese bert-networks,'' in \emph{Conference on Empirical Methods in Natural Language Processing}, 2019.

\bibitem{lewis2020bart}
M.~Lewis, Y.~Liu, N.~Goyal, M.~Ghazvininejad, A.~Mohamed, O.~Levy, V.~Stoyanov, and L.~Zettlemoyer, ``Bart: Denoising sequence-to-sequence pre-training for natural language generation, translation, and comprehension,'' in \emph{Proceedings of the 58th Annual Meeting of the Association for Computational Linguistics (ACL)}, 2020, pp. 7871--7880.

\bibitem{young2014image}
P.~Young, A.~Lai, M.~Hodosh, and J.~Hockenmaier, ``From image descriptions to visual denotations: New similarity metrics for semantic inference over event descriptions,'' \emph{Transactions of the Association for Computational Linguistics}, vol.~2, pp. 67--78, 2014.

\end{thebibliography}


\begin{IEEEbiography}[{\includegraphics[width=1in,height=1.25in,clip,keepaspectratio]{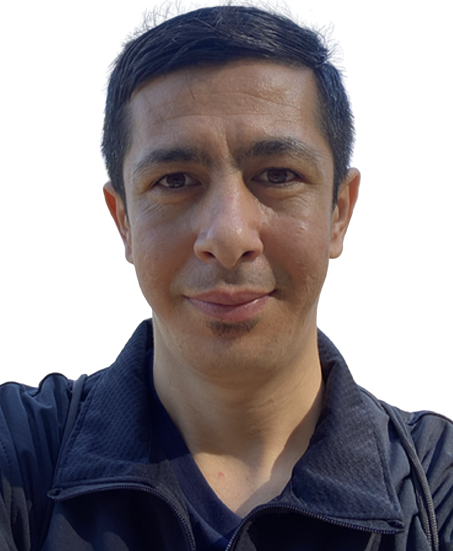}}]{Javad Rafiei Asl} received his B.S. degree in computer science from the University of Tabriz, Iran, in 2011; the M.S. degree in information technology from the University of Tehran , Iran, in 2014. He then joined the Research Institute for Information and Communication Technology (ICTRC), Tehran, Iran as a research assistant for developing an automated plagiarism detection system analyzing scientific articles and dissertations. After achieving practical experience in data mining and machine learning, he joined Georgia State University, in 2019, to pursue his Ph.D. education as a computer scientist in Natural Language Processing (NLP) and Machine Learning (ML) domains. Javad is currently a researcher member of the Information Security and Privacy: Interdisciplinary Research and Education (INSPIRE) Center at Georgia State University, working on the cutting-edge domain of secure artificial intelligence for different NLP tasks.
\end{IEEEbiography}

\begin{IEEEbiography}[{\includegraphics[width=1in,height=1.25in,clip,keepaspectratio]{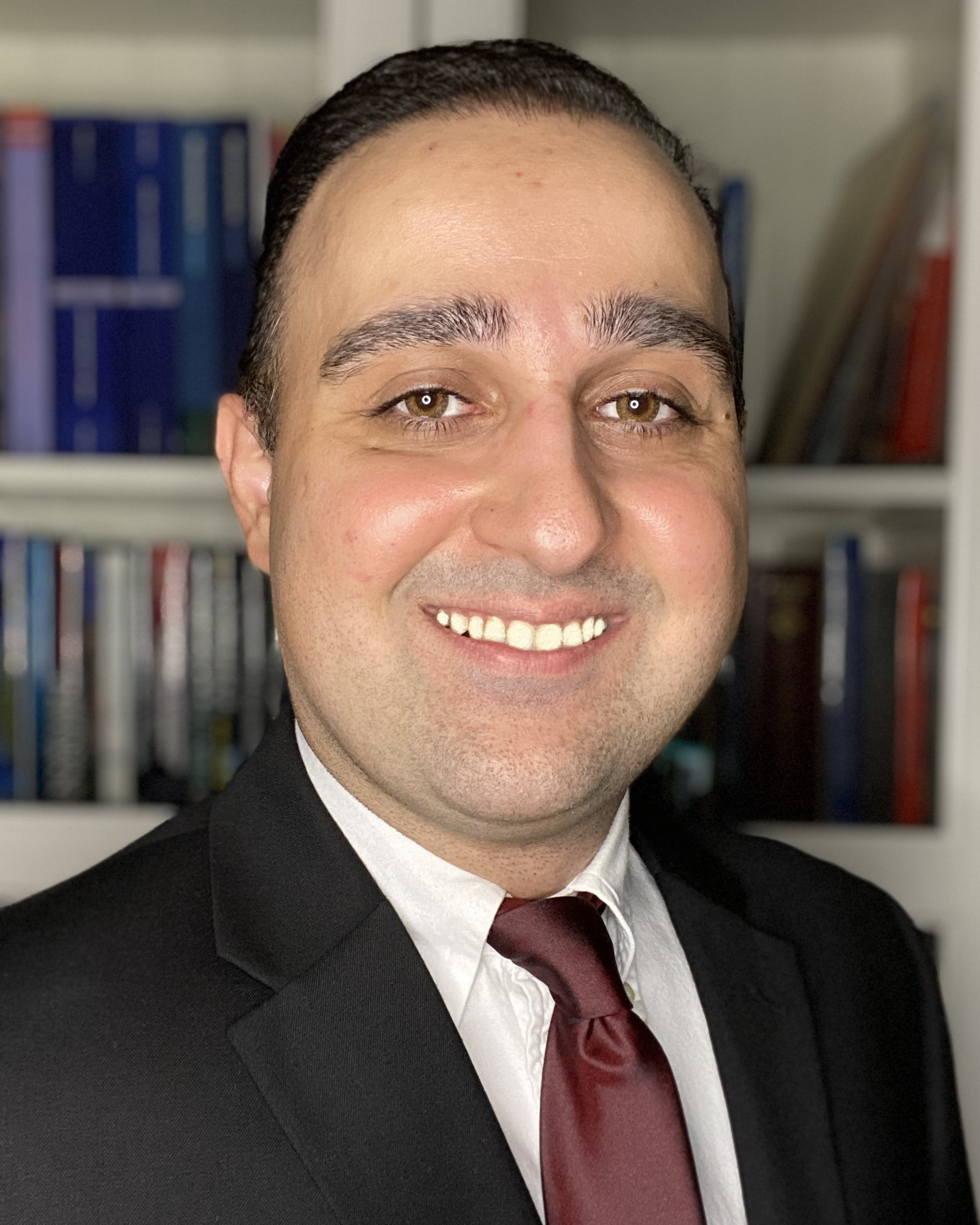}}]{Mohammad H. Rafiei} obtained his Ph.D. in Civil Engineering focused on Machine Learning from Ohio State University in Dec 2016 under Professor Hojjat Adeli. Dr. Rafiei continued his Machine Learning research as a Postdoctoral Researcher at Ohio State University Engineering and Medical Schools between Feb 2017 and Aug 2018 in Infrastructure Engineering, Neuroscience, and Physical Medicine. He then joined Johns Hopkins University Whiting School of Engineering as a Postdoctoral Fellow in Aug 2018 and an Adjunct Assistant Research Scientist in Sep 2019, working in Machine Learning Nano-Scale Materials Simulations. He served as a Machine Learning course developer and instructor at the Engineering for Professionals graduate-level program at Johns Hopkins University, Whiting School of Engineering, in 2020. Dr. Rafiei joined the Information Security and Privacy: Interdisciplinary Research and Education (INSPIRE) Center at Georgia State University, Department of Computer Science in Jan 2021 as a Postdoctoral Fellow working on the emerging area of adversarial machine learning.
\end{IEEEbiography}

\begin{IEEEbiography}[{\includegraphics[width=1in,height=1.25in,clip,keepaspectratio]{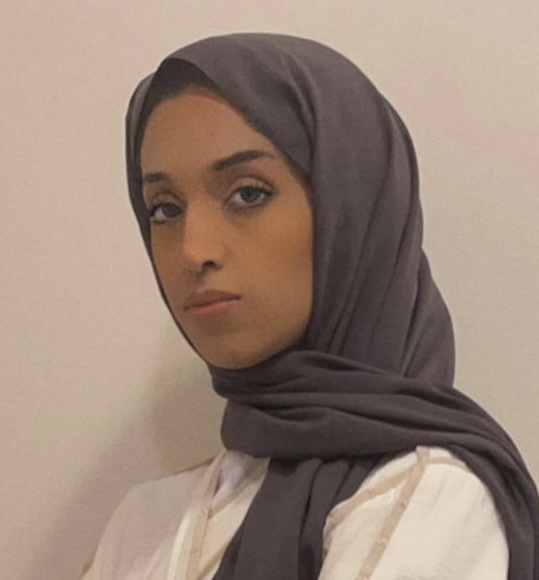}}]{Manar Alohaly} is an Assistant Professor of Cybersecurity at Princess Nourah Bint Abdulrahman University (PNU), Riyadh, Saudi Arabia. She is currently the director of the Innovation Center at the College of Computer and Information Sciences, PNU. In 2020, she received her Ph.D. from the University of North Texas, Denton, Texas, USA. Her research interests include access control, usable privacy and security, cyber deception, insider threat detection, applied machine learning, and natural language processing on security-related topics. She is a member of the technical program committee for several international conferences.
\end{IEEEbiography}

\begin{IEEEbiography}[{\includegraphics[width=1in,height=1.25in,clip,keepaspectratio]{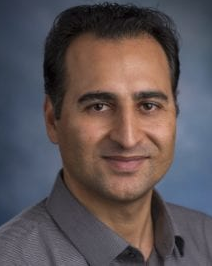}}]{Daniel Takabi} (Member, IEEE) received the Ph.D. degree from the University of Pittsburgh, Pittsburgh, PA, USA, in 2013. He is currently an Associate Professor of computer science and the Next Generation Scholar with Georgia State University (GSU), Atlanta, GA, USA. He is also a Founding Director of the Information Security and Privacy: Interdisciplinary Research and Education (INSPIRE) Center, which is designated as the National Center of Academic Excellence in Cyber Defense Research (CAE-R). His research interests include various aspects of cybersecurity and privacy, including privacy preserving machine learning, adversarial machine learning, advanced access control models, insider threats, and usable security and privacy. 
\end{IEEEbiography}

% \begin{IEEEbiography}[{\includegraphics[width=1in,height=1.25in,clip,keepaspectratio]{authors/Takabi}}]{Daniel Takabi} is currently an Associate Professor of computer science and the Next Generation Scholar with Georgia State University. He is also a Founding Director of the Information Security and Privacy: Interdisciplinary Research and Education (INSPIRE) Center, designated as the National Center of Academic Excellence in Cyber Defense Research (CAE-R). His research interests include various aspects of cybersecurity and privacy, including trustworthy AI, privacy-preserving machine learning, adversarial learning, advanced access control models, insider threats, and usable security and privacy.
% \end{IEEEbiography}


%\vfill


\end{document}